\documentclass[letterpaper]{article} 
\usepackage{aaai2027}  
\providecommand{\CogVisHeavyCheck}{\ensuremath{\checkmark}}
\usepackage[hyphens]{url}  
\usepackage{graphicx} 
\usepackage{natbib}  
\usepackage{caption} 
\usepackage{amsmath,amssymb}
\usepackage{array}
\usepackage{xcolor}
\usepackage{colortbl}
\usepackage{multirow}
\usepackage{algorithm}
\usepackage{algorithmic}

\providecommand{\CogVisDoubleSep}{\vrule width 0.30pt\hskip0.65pt\vrule width 0.30pt}
\providecommand{\CogVisTableTopRule}{\toprule[0.85pt]}
\providecommand{\CogVisTableHeaderRule}{\midrule[0.60pt]\addlinespace[0.08em]}
\providecommand{\CogVisTableGroupRule}{\midrule[0.35pt]\addlinespace[0.12em]}
\providecommand{\CogVisTableSummaryRule}{\midrule[0.50pt]\addlinespace[0.10em]}
\providecommand{\CogVisTableBottomRule}{\bottomrule[0.85pt]}

\providecommand{\CogVisOursCell}[1]{\textbf{#1}}
\providecommand{\CogVisBestCell}[1]{\textbf{#1}}
\providecommand{\CogVisSecondCell}[1]{\underline{#1}}
\providecommand{\CogVisAblMark}{\CogVisHeavyCheck}
\providecommand{\CogVisAblDash}{--}
\usepackage{booktabs}

\title{CogVis: Must Open-Vocabulary Change Detection Perceive the Scene Anew for Every Query?}
\author{
    Zijie Wang,
    Chen Zhong,
    Wei He\corresponding
}
\affiliations{
    Key Laboratory of Information Engineering in Surveying, Mapping and Remote Sensing, Wuhan University\\
    Wuhan, China\\
    zijie.wang@whu.edu.cn, chen\_zhong@whu.edu.cn, weihe1990@whu.edu.cn
}

\newcolumntype{L}[1]{>{\raggedright\arraybackslash}p{#1}}
\newcolumntype{C}[1]{>{\centering\arraybackslash}p{#1}}

\newcommand{\CogVisCompactTableSetup}{%
  \footnotesize
  \setlength{\tabcolsep}{0.75mm}%
  \renewcommand{\arraystretch}{1.00}%
}
\newcommand{\CogVisPromptTableSetup}{%
  \footnotesize
  \setlength{\tabcolsep}{0.75mm}%
  \renewcommand{\arraystretch}{1.00}%
}
\newcommand{\clip}{\operatorname{clip}}
\makeatletter
\let\CogVisSavedTitle\@title
\newcommand{\CogVisSupplementaryTitle}{%
  \par
  \twocolumn[{%
    \vbox to 1.38in{%
      \vskip 0.18in
      \centering
      {\LARGE\bfseries \CogVisSavedTitle\par}%
      \vskip 0.10in
      {\Large\bfseries Supplementary Material\par}%
      \vfill
    }%
  }]%
  \thispagestyle{empty}%
}
\makeatother

\nocopyright

\begin{document}

\maketitle

\begin{abstract}
Earth-surface monitoring requires change detection models capable of recognizing arbitrary semantic categories. Open-Vocabulary Change Detection (OVCD) addresses this need. However, existing methods often entangle temporal perception, semantic discrimination, and region verification, causing unstable results and redundant computation. Inspired by human visual change perception, we propose CogVis, a cognitive memory-guided framework that reformulates OVCD as a perception-memory-verification paradigm. CogVis first employs a Scene Change Perceptron (SCP) to extract a reusable, category-agnostic change prior from frozen bi-temporal features, thereby decoupling temporal evidence from semantic category decisions. A Semantic Memory Calibrator (SMC) then compensates for category-dependent score shifts by dynamically estimating an image-query-specific decision threshold. Finally, an Adaptive Region Filter (ARF) filters connected candidates using learned semantic, temporal, and structural reliability. Experiments on seven benchmarks spanning semantic change detection, binary change localization, and building-damage assessment show that CogVis achieves state-of-the-art performance across all evaluated datasets. By sharing scene-level change perception, CogVis further avoids repeating category-agnostic temporal perception across queries and improves inference throughput by 28.50\%.

\end{abstract}

\begin{links}
    \link{Code}{https://github.com/KotlinWang/CogVis}
\end{links}

\section{Introduction}
Earth’s surface is continuously reshaped by natural processes and human activities, producing complex spatiotemporal patterns that must be reliably interpreted for environmental monitoring and informed decision-making \cite{li2026perspectives}. Change Detection (CD) identifies changes in multitemporal remote sensing imagery and supports applications including land-cover monitoring, disaster assessment, urban planning, and ecological observation \cite{hong2024spectralgpt,wang2025accurate}. However, most CD research assumes that scene distributions and semantic categories are fixed during training. This closed-set assumption limits their applicability in open-world scenarios, where acquisition conditions vary and users may query categories that are absent from the task-specific classifier. \par
Vision models and Vision-Language Models (VLMs) provide transferable priors for segmentation, representation learning, and text-image alignment. SAM3 \cite{carion2025sam3}, DINOv3 \cite{simeoni2025dinov3}, and CLIP \cite{radford2021clip} have consequently enabled zero-shot or training-light remote sensing interpretation. Foundation-model-based change detectors such as AnyChange \cite{zheng2024anychange} and UCD-SCM \cite{tan2024scm} improve category-agnostic change localization but primarily determine whether a region has changed, not which text-specified category accounts for the change. \par

\begin{figure}[t]
    \centering
    \includegraphics[width=\linewidth]{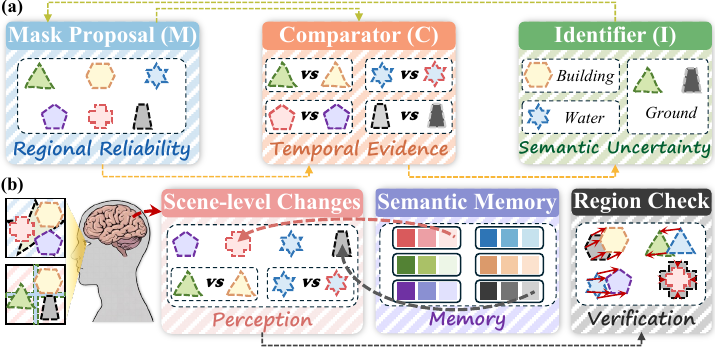}
    \caption{Comparison of OVCD paradigms. Existing M-C-I and I-M-C pipelines couple temporal, semantic, and regional decisions, whereas CogVis separates them into shared perception, memory-guided calibration, and verification.}
    \label{fig_1}
\end{figure}

Open-Vocabulary Change Detection (OVCD) has emerged to bridge this gap by detecting bi-temporal changes under text-specified concepts. DynamicEarth \cite{li2026dynamicearth} summarizes existing research through the Mask-Compare-Identify (M-C-I) and Identify-Mask-Compare (I-M-C) pipelines. M-C-I shares category-independent proposals, but downstream recognition inherits their localization errors. Missed candidates cannot be recovered, whereas spurious candidates may still receive plausible labels \cite{dou2026adaptovcd,guo2026opendpr}. I-M-C injects category priors before temporal comparison, enabling category-aware reasoning but repeatedly parses the image pair for every query \cite{zhu2025semantic,zhang2026omniovcd,tang2026coregovcd,kuang2026memovcd}. Therefore, both designs couple temporal evidence, semantic uncertainty, and regional reliability within a fixed cascade, as shown in Fig.~\ref{fig_1}(a). \par
Unlike the fixed logic of M-C-I or I-M-C, we formulate OVCD as a perception-memory-verification paradigm motivated by human visual change perception. As shown in Fig.~\ref{fig_1}(b), humans first form a global perception of scene-level changes, then leverage semantic memory to infer plausible categories, and finally verify each candidate region to reject non-semantic disturbances such as shadows, texture variations, seasonal shifts, and misregistration \cite{lerebourg2024attentional, brady2024noisy}. Based on this observation, we propose CogVis, a cognitive memory-guided OVCD framework that follows a scene-first and memory-calibrated process. CogVis first extracts a reusable category-agnostic temporal prior with a Scene Change Perceptron (SCP). A Semantic Memory Calibrator (SMC) then contrasts each query with nearby semantic confusers and retrieves memory cases to estimate an image-query-specific decision threshold. Finally, an Adaptive Region Filter (ARF) verifies connected candidates through semantic, temporal, and structural agreement. The main contributions of this work are as follows:
\begin{itemize}
    \item Inspired by human visual change perception, we reformulate OVCD as a perception-memory-verification paradigm that disentangles the decisions coupled within the fixed M-C-I and I-M-C cascades.
    \item We instantiate this paradigm as CogVis, where SCP, SMC, and ARF jointly address semantic-temporal-structural entanglement, image-query score shifts, and unreliable region predictions.
    \item We validate this factorization on seven benchmarks, demonstrating consistent cross-task gains and reduced multi-query temporal computation.
\end{itemize}

\section{Related Work}

\subsection{Foundation Models for Remote Sensing}
Foundation models are increasingly replacing task-specific representation learning in remote sensing, reflecting a broader shift toward universal large-scale remote sensing models \cite{zhang2026earthscale}. CLIP aligns imagery with open-vocabulary concepts \cite{radford2021clip,liu2024remoteclip}, DINO-style encoders provide spatially coherent dense features \cite{caron2021emerging,oquab2024dinov2}, and SAM-style models generate promptable masks \cite{kirillov2023segment,carion2025sam3}. Methods such as SegEarth-OV, ReSeg-CLIP, and SeeCo further adapt these priors to remote-sensing challenges including scale variation, weak boundaries, and domain-specific semantics \cite{li2025segearthov,heidarianbaei2026resegclip,wang2026seeco}. However, these advances primarily target single-image interpretation. Change understanding additionally requires temporally comparable evidence while preserving open-vocabulary semantics. Therefore, CogVis retains the frozen semantic pathway and learns only lightweight adapters for temporal perception, calibration, and verification.

\subsection{Open-Vocabulary Change Detection}
OVCD extends CD from fixed categories to text-specified semantic changes. DynamicEarth formalizes the dominant M-C-I and I-M-C paradigms \cite{li2026dynamicearth}. AdaptOVCD aligns heterogeneous model outputs through adaptive fusion \cite{dou2026adaptovcd}, OpenDPR retrieves visual prototypes for open-category recognition \cite{guo2026opendpr}, Seg2Change transfers open-vocabulary segmentation to CD \cite{su2026seg2change}, and OmniOVCD addresses semantic and instance mismatches \cite{zhang2026omniovcd}. Although these methods improve individual stages, they do not explicitly separate query-shared temporal perception, query-specific boundary calibration, and region-level verification. CogVis instead organizes these decisions according to their scope rather than embedding them in a fixed processing cascade.

\subsection{Memory for Change Reasoning}
In conventional CD, memory typically stores temporal states or feature prototypes. MS-Former maintains changed and unchanged prototypes, whereas ChangeTitans employs neural memory for long-range temporal modeling \cite{li2024msformer,yang2025changetitans}. MemOVCD further uses memory for cross-temporal semantic propagation and global-local rectification \cite{kuang2026memovcd}. CogVis adopts a fundamentally different role for memory. Each entry associates a continuous scene-temporal semantic response state with a calibrated threshold. Therefore, retrieval transfers an operating point rather than a feature prototype, which the Score Adapter further adjusts for the current image-query pair. This design directly compensates for scene- and category-dependent score shifts without modifying the semantic response map.

\begin{figure*}[t]
    \centering
    \includegraphics[width=\linewidth]{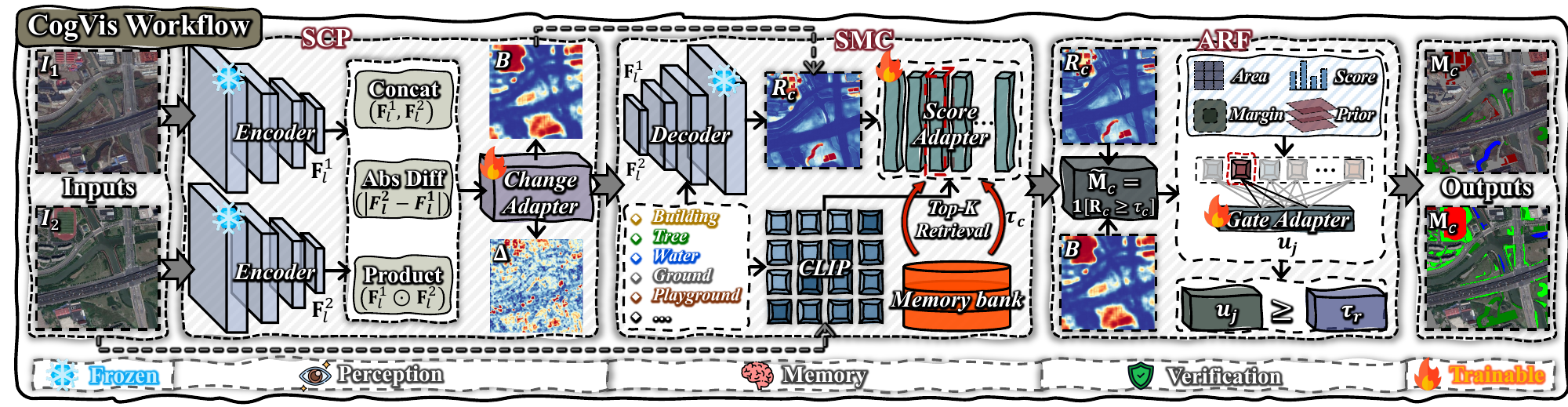}
    \caption{Overview of CogVis. SCP extracts a query-shared change prior, SMC calibrates query-specific responses with memory, and ARF verifies candidate regions.}
    \label{fig_2}
\end{figure*}

\section{Methodology}

\textbf{\textit{Problem Formulation:}} Given bi-temporal remote sensing images $I^{1},I^{2}\in\mathbb{R}^{H\times W\times 3}$ and the open vocabulary $\mathcal{C}=\{c_1,\ldots,c_K\}$, OVCD predicts a change mask for each query
\begin{equation}
\mathcal{M}
=
\left\{
\mathbf{M}_c\in\{0,1\}^{H\times W}
\mid c\in\mathcal{C}
\right\}.
\label{eq:ovcd_problem}
\end{equation}
The task contains three decisions with different scopes. Temporal evidence is shared across queries, semantic confidence depends on the image-query pair, and candidate regions require reliability verification. CogVis addresses them with SCP, SMC, and ARF, respectively. \par
\textbf{\textit{Framework Overview:}} As shown in Fig.~\ref{fig_2}, a frozen visual encoder extracts bi-temporal features. SCP compares the image pair once to produce a reusable change prior $\mathbf{B}$. For each query $c$, SMC decodes the original frozen features, constructs a semantic-change response $\mathbf{R}_c$, and estimates an adaptive threshold $\tau_c$. ARF then verifies connected candidates to obtain $\mathbf{M}_c$. The SCP residual is used only for $\mathbf{B}$ and never alters semantic decoding.

\subsection{Scene Change Perceptron}
SCP extracts query-shared temporal evidence once per image pair. Let
$\{\mathbf{F}_l^t\}_{l=1}^{L}=\mathcal{E}(I^t)$ denote the frozen SAM3
feature pyramid at time $t\in\{1,2\}$. We denote the second-level features
by $\mathbf{F}^{t}=\mathbf{F}_{2}^{t}$ and construct
\begin{equation}
\mathbf{X}
=
\operatorname{Cat}_{\mathrm{ch}}\!\left(
\mathbf{F}^{1},
\mathbf{F}^{2},
|\mathbf{F}^{2}-\mathbf{F}^{1}|,
\mathbf{F}^{1}\odot\mathbf{F}^{2}
\right).
\label{eq:scp_descriptor}
\end{equation}
A Change Adapter maps $\mathbf{X}$ to a residual
$\widehat{\boldsymbol{\Delta}}
=\mathbf{W}_{\uparrow}\phi(\mathbf{W}_{\downarrow}\mathbf{X})$.
A temporal gate produces $\boldsymbol{\Delta}$ by capping its root-mean-square magnitude
at $25\%$ of the frozen-feature magnitude. The change head then predicts
\begin{equation}
\mathbf{B}
=
\sigma\!\left(
\operatorname{Up}\bigl(
\mathcal{H}_{\mathrm{chg}}(\mathbf{X},\boldsymbol{\Delta})
\bigr)
\right).
\label{eq:scp_prior}
\end{equation}
The resulting category-agnostic prior is shared by all queries, while the
original feature pyramid is preserved for semantic decoding.

\subsection{Semantic Memory Calibrator}
Open-vocabulary responses vary across scenes and categories, making a fixed decision boundary unreliable. Therefore, SMC constructs a query-specific semantic-change response and calibrates its threshold from memory. \par
\textbf{\textit{{Semantic-change Response:}}} For query $c$, the frozen decoder processes prompt set $\mathcal{P}_c$ using the original temporal features. Let $\mathbf{Q}_c^t$ denote the aggregated target response at time $t$, and $\mathbf{C}_c^t$ the strongest response among CLIP-selected semantic neighbors. Prompt competition suppresses ambiguous activation as
\begin{equation}
\widetilde{\mathbf{Q}}_c^t
=
\mathbf{Q}_c^t\odot
\left(
\frac{\mathbf{Q}_c^t}
{\mathbf{Q}_c^t+\mathbf{C}_c^t+\epsilon}
\right)^{\rho},
\quad t\in\{1,2\},
\label{eq:smc_competition}
\end{equation}
where $\rho$ controls competition strength. The competition-adjusted responses produce four cues. $\mathbf{D}_c$ captures semantic transitions, $\mathbf{G}_c$ enforces support from the change prior $\mathbf{B}$, $\mathbf{S}_c$ suppresses persistent semantics in stable regions, and $p_c$ measures the change in query presence. Their agreement defines
\begin{equation}
\mathbf{R}_c
=
\Pi_{[0,1]}\!\left(
\mathbf{D}_c\odot\mathbf{G}_c\odot\mathbf{S}_c\,p_c
\right),
\label{eq:smc_response}
\end{equation}
The resulting response retains query-specific transitions supported by temporal evidence. \par
\textbf{\textit{{Memory Construction and Retrieval:}}} Each source image-query pair stores a normalized $1556$-dimensional key and its oracle threshold. The key concatenates scene context, temporal discrepancy, and query embeddings in $\mathbb{R}^{512}$ with $20$ statistics describing $\mathbf{R}_c$ and its agreement with $\mathbf{B}$. The oracle threshold maximizes IoU between the thresholded response and the query-specific source mask over a predefined grid. \par
At inference, cosine similarity retrieves the top $K_{\mathrm{mem}}=16$ memory entries. Their thresholds form a temperature-weighted anchor
\begin{equation}
\pi_i^c
=
\frac{\exp(s_i/T_m)}
{\sum_{j\in\mathcal{K}_c}\exp(s_j/T_m)},
\qquad
\bar{\tau}_c
=
\sum_{i\in\mathcal{K}_c}\pi_i^c\tau_i^{\star},
\label{eq:memory_retrieval}
\end{equation}
where $s_i^{c}=\mathbf{k}_c^{\top}\mathbf{k}_i$ and $T_m=0.07$. A Score Adapter predicts a bounded correction $\Delta\tau_c$ from the response statistics, retrieved anchor, and similarity confidence. Combined with the label-free threshold $\tau_c^{0}=\mathcal{T}_{\mathrm{adp}}(\mathbf{R}_c)$, the boundary is
\begin{equation}
\tau_c
=
\Pi_{[\tau_{\min},\tau_{\max}]}\!\left(
\max\{\bar{\tau}_c+\Delta\tau_c,\,0.75\tau_c^{0}\}
\right).
\label{eq:smc_threshold}
\end{equation}
Here, $\Pi_{[a,b]}(\cdot)$ denotes element-wise clipping to $[a,b]$. The retrieved anchor transfers a suitable operating point, while the learned correction adapts it to the current image-query pair. Thresholding $\mathbf{R}_c$ at $\tau_c$ yields the candidate mask $\widetilde{\mathbf{M}}_c$.

\begin{table*}[t]\centering
\small
\setlength{\tabcolsep}{1mm}
\providecommand{\CogVisGroupRow}[1]{\rowcolor[gray]{0.94}\multicolumn{10}{@{}l@{}}{\textbf{\textit{#1}}} \\}
\begin{tabular}{@{\hspace{\tabcolsep}}cccc!{\CogVisDoubleSep}cccc!{\CogVisDoubleSep}c!{\CogVisDoubleSep}c@{\hspace{\tabcolsep}}}
\CogVisTableTopRule
\multirow[c]{2}{*}{\textbf{Method}} & \multirow[c]{2}{*}{\textbf{Venue}} &
\multicolumn{2}{c}{\textbf{Semantic CD (\%)}} &
\multicolumn{4}{c}{\textbf{Binary CD (\%)}} &
\multicolumn{1}{c}{\textbf{Disaster (\%)}} &
\multirow[c]{2}{*}{\textbf{\#P $\downarrow$}} \\
\cmidrule(lr){3-4}\cmidrule(lr){5-8}\cmidrule(lr){9-9}
 & & \textbf{SECOND} & \textbf{SCSCD} & \textbf{CLCD} & \textbf{DSIFN} & \textbf{LEVIR-CD} & \textbf{WHU-CD} & \textbf{xBD} & \\
\CogVisTableHeaderRule
\CogVisGroupRow{Category-agnostic CD}
PCA-KMeans & GRSL09 & $-$ & $-$ & 10.14 & 21.57 & 5.24 & 7.72 & $-$ & $-$ \\
CVA & RSE94 & $-$ & $-$ & 7.17 & 16.20 & 4.86 & 3.72 & $-$ & $-$ \\
DCVA & TGRS19 & $-$ & $-$ & 11.62 & 23.85 & 7.44 & 11.45 & $-$ & 10 \\
UCD-SCM & IGARSS24 & 8.40 & 6.51 & 13.19 & 25.10 & 19.30 & 19.14 & 19.04 & 219 \\
AnyChange & NeurIPS24 & 11.84 & 9.32 & 19.02 & 24.37 & 19.53 & 16.37 & 19.63 & 641.09 \\
Inst-CEG & TGRS25 & 10.46 & 3.68 & 3.50 & 18.91 & 46.30 & 45.49 & $-$ & $-$ \\
\CogVisTableGroupRule
\CogVisGroupRow{Cascaded OVCD}
DynamicEarth-MCI & AAAI26 & 23.58 & 17.97 & 13.52 & 37.32 & 30.23 & 40.20 & \underline{24.51} & 1087 \\
DynamicEarth-IMC & AAAI26 & 13.72 & 5.16 & 8.09 & 15.22 & 53.50 & 61.09 & $-$ & 584 \\
UniVCD & arXiv25 & 20.53 & $-$ & $-$ & $-$ & 56.70 & 66.00 & $-$ & $-$ \\
SAM3-CD & arXiv25 & 16.12 & $-$ & $-$ & $-$ & 40.00 & 36.30 & $-$ & $-$ \\
AdaptOVCD & arXiv26 & 28.05 & $-$ & $-$ & 42.32 & 51.52 & 61.99 & 22.07 & 1393.78 \\
OpenDPR & CVPR26 & 28.55 & $-$ & $-$ & $-$ & 44.80 & 54.30 & 19.58 & 727.67 \\
OmniOVCD & arXiv26 & 27.05 & $-$ & $-$ & 24.50 & 67.20 & 66.50 & 18.92 & 840.51 \\
CoRegOVCD & arXiv26 & \underline{31.67} & $-$ & $-$ & \underline{47.55} & 70.95 & 69.70 & $-$ & $-$ \\
Seg2Change & arXiv26 & 29.08 & \underline{23.22} & \underline{31.48} & 41.40 & 64.91 & \underline{75.72} & 17.94 & 931.01 \\
\CogVisTableGroupRule
\CogVisGroupRow{Memory-augmented OVCD}
MemOVCD & arXiv26 & $-$ & $-$ & $-$ & 37.80 & \underline{72.50} & $-$ & $-$ & $-$ \\
CogVis (Ours) & -- & \CogVisOursCell{32.48} & \CogVisOursCell{25.36} & \CogVisOursCell{33.67} & \CogVisOursCell{48.50} & \CogVisOursCell{72.55} & \CogVisOursCell{75.86} & \CogVisOursCell{32.68} & 990.29 \\ \CogVisTableSummaryRule 	
\rowcolor[gray]{0.94}\textbf{Margin} & -- & +0.81 & +2.14 & +2.19 & +0.95 & +0.05 & +0.14 & +8.17 & $-$ \\
\CogVisTableBottomRule
\end{tabular}
\normalsize
\caption{Comparison on seven benchmarks. We report mIoU for semantic CD and xBD and IoU for binary CD. Bold and underlined indicate the best and second-best results, respectively; ``$-$'' denotes an unreported result. ``Margin'' is the gain over the strongest prior result, and \#P denotes total parameters (M).}
\label{tab:ovcd_seven_dataset_comparison}
\end{table*}

\subsection{Adaptive Region Filter}
Pixelwise thresholding may retain fragmented or weak responses. Therefore, ARF verifies connected regions before prediction. Let $\Omega_c=\operatorname{CCL}(\widetilde{\mathbf{M}}_c)$ denote the candidate components. The CCL denotes connected-component labeling. After deterministic component and area handling, each $\omega_j\in\Omega_c$ is encoded by a $19$-dimensional descriptor $\boldsymbol{\varphi}_j$ covering geometry, semantic confidence, temporal support, calibrated margin, and semantic-temporal agreement. The Gate Adapter predicts its reliability as $u_j=\sigma(\mathcal{A}_{\mathrm{gate}}(\boldsymbol{\varphi}_j))$. \par
Let $a_j=|\omega_j|/(HW)$ denote the area ratio of component $\omega_j$. We set the large-region protection threshold to $\eta_a=0.04$ and the Gate threshold to $\tau_r=0.60$. ARF retains a component if it satisfies either criterion:
\begin{equation}
\mathbf{M}_c
=
\bigcup
\left\{
\omega_j\in\Omega_c
\;\middle|\;
a_j\geq\eta_a
\;\lor\;
u_j\geq\tau_r
\right\}.
\label{eq:arf_decision}
\end{equation}
This rule preserves large components directly, while smaller candidates must pass learned semantic-temporal-structural verification. ARF removes unreliable proposals but cannot recover regions missing from $\widetilde{\mathbf{M}}_c$.

\subsection{Adapter Training}
The visual encoder, semantic decoder, and CLIP encoders remain frozen. The Change, Score, and Gate Adapters are optimized sequentially as separate checkpoints. \par
The Change Adapter is trained with the binary-change objective $\mathcal{L}_{\mathrm{chg}}$, comprising equally weighted balanced BCE-with-logits and soft Dice losses, to produce the prior in Eq.~\ref{eq:scp_prior}. Subsequently, the source responses are generated, and the calibration memory is constructed offline. For source entry $i$, leave-one-out retrieval yields $\bar{\tau}_i^{(-i)}$, and the Score Adapter regresses the residual to the oracle threshold using Smooth-$L_1$ loss:
\begin{equation}
\mathcal{L}_{\mathrm{score}}
=
\frac{1}{N}\sum_{i=1}^{N}
\ell_{\mathrm{sl1}}\!\left(
\Delta\tau_i,
\tau_i^{\star}-\bar{\tau}_i^{(-i)}
\right).
\label{eq:score_loss}
\end{equation}
Finally, connected candidates are labeled by the source-overlap rule, and the Gate Adapter is trained with binary cross-entropy:
\begin{equation}
\mathcal{L}_{\mathrm{gate}}
=
\frac{1}{N_{\Omega}}
\sum_{j=1}^{N_{\Omega}}
\ell_{\mathrm{bce}}(u_j,y_j),
\label{eq:gate_loss}
\end{equation}
where $y_j\in\{0,1\}$ is the target reliability of $\omega_j$. This staged procedure preserves the role of each adapter and avoids coupling the three decision stages during optimization.

\begin{figure*}[t]
    \centering
    \includegraphics[width=\linewidth]{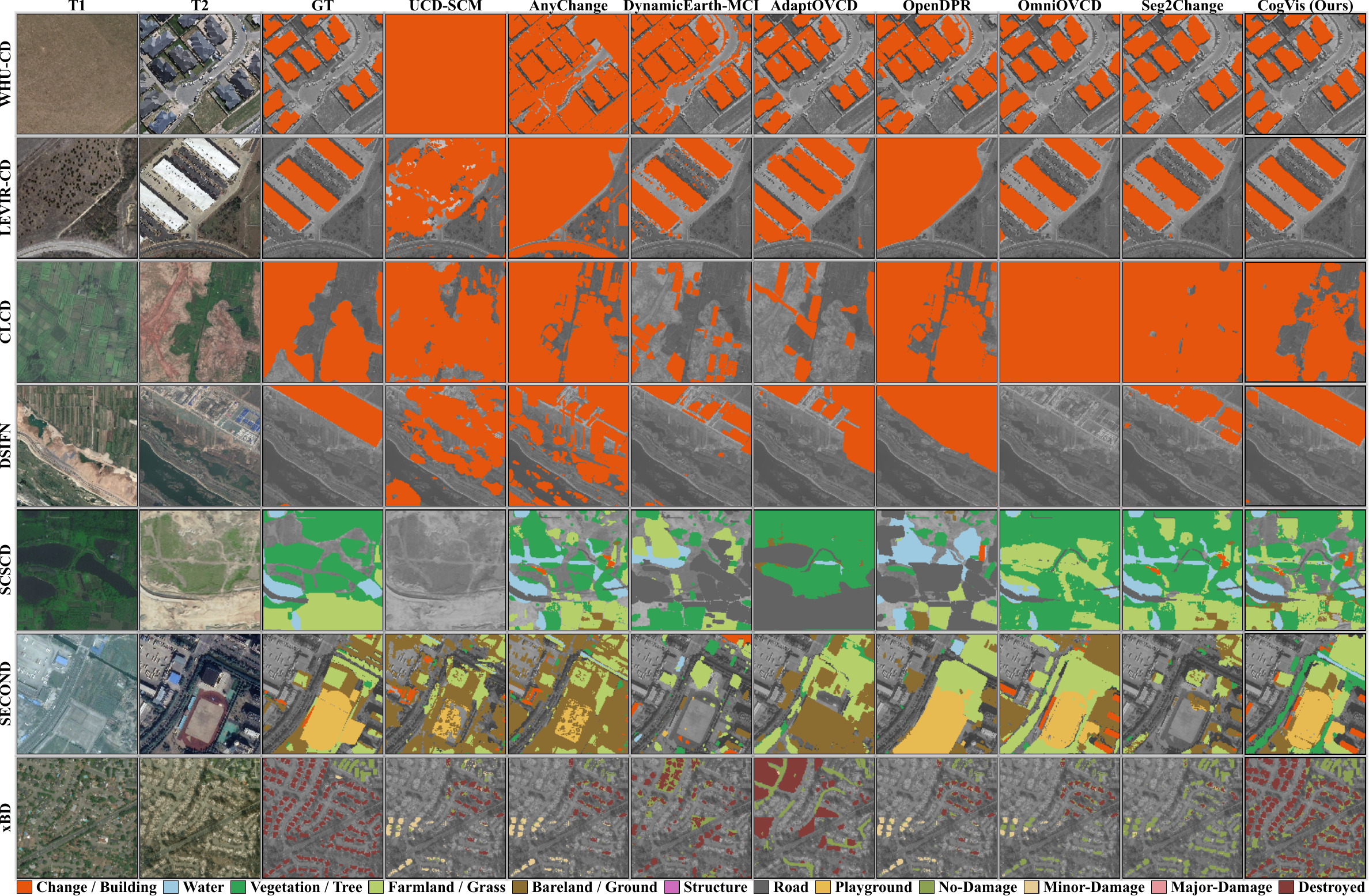}
    \caption{Qualitative comparison across seven benchmarks covering semantic CD, binary CD, and building-damage assessment. Ground truth and predictions follow the class legend.}
    \label{fig_3}
\end{figure*}

\begin{figure}[t]
    \centering
    \includegraphics[width=\linewidth]{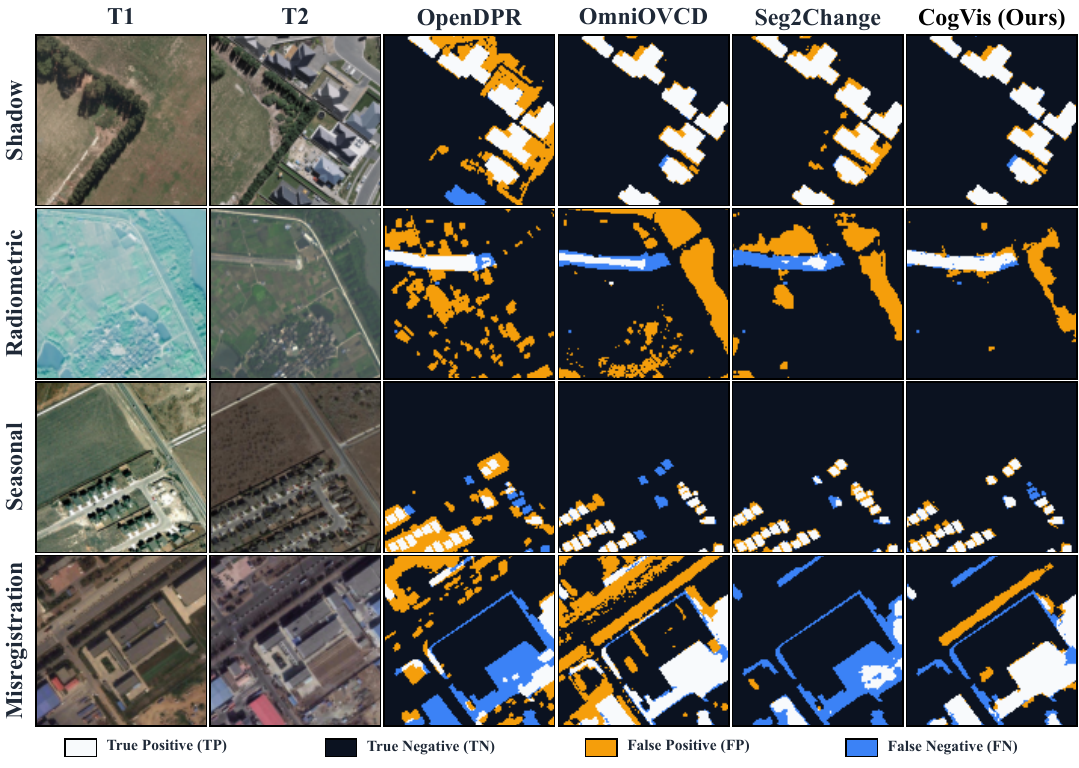}
    \caption{Qualitative comparison across four nuisance conditions: shadows, radiometric or textural variations, seasonal shifts, and residual misregistration.}
    \label{fig_4}
\end{figure}

\begin{figure*}[t]
    \centering
    \includegraphics[width=\linewidth]{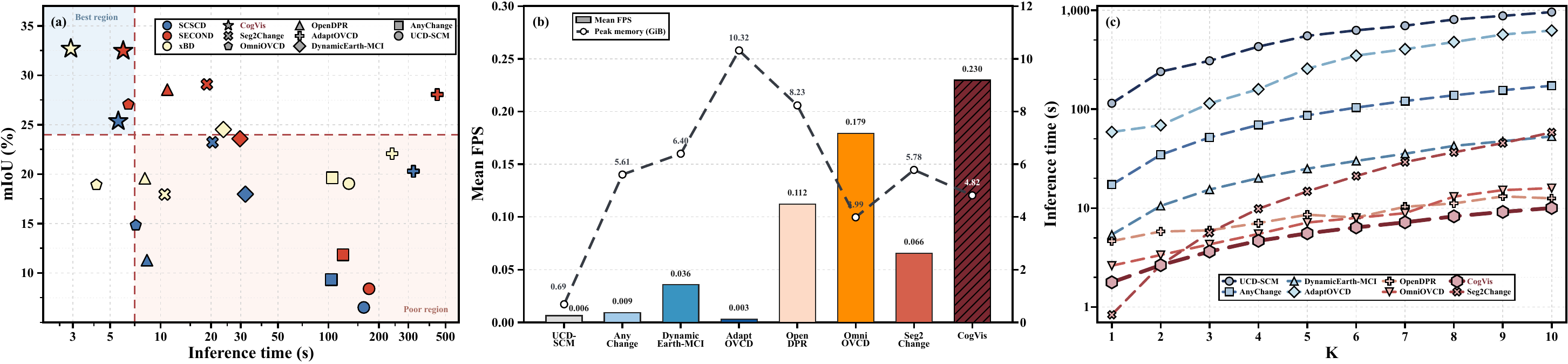}
    \caption{Accuracy-efficiency and query-scaling analysis on RTX 3090 GPU with $512\times512$ inputs. (a) Inference time versus mIoU. (b) Mean throughput and peak memory. (c) Latency as the number of queried categories increases.}
    \label{fig_5}
\end{figure*}

\section{Experiments and Analysis}

\subsection{Experimental Settings}
\textbf{\textit{{Benchmarks:}}} We evaluate CogVis on seven public benchmarks covering semantic CD, including SECOND \cite{yang2022second} and SCSCD \cite{tan2025triples}; binary CD, including CLCD \cite{liu2022clcd}, DSIFN \cite{zhang2020dsifn}, LEVIR-CD \cite{chen2020levir}, and WHU-CD \cite{ji2019whu}; and building-damage assessment on xBD \cite{gupta2019xbd}. CA-CDD \cite{su2026seg2change} is used exclusively for adapter optimization and memory construction and does not overlap with any test set. \par
\textbf{\textit{{Metrics:}}} We report mIoU as the primary metric for semantic CD and building-damage assessment, and IoU for binary CD. Efficiency is measured by latency, frames per second (FPS), peak GPU memory, and parameter count. Query-scaling experiments use $512\times512$ inputs and are conducted on an RTX 3090 GPU. \par
\textbf{\textit{{Implementation:}}} CogVis uses frozen SAM3 \cite{carion2025sam3} visual features and CLIP ViT-B/16 \cite{radford2021clip} embeddings for text encoding and retrieval. The Change, Score, and Gate Adapters are optimized sequentially on CA-CDD for 100 epochs with a batch size of 16. The top 16 memory-bank entries are retrieved with a temperature of $0.07$. The ARF gating and large-region thresholds are set to $0.60$ and $0.04$, respectively.

\subsection{Comparison with State-of-the-Art Methods}
\textbf{\textit{{Quantitative Comparison:}}} CogVis achieves the best metric on all seven benchmarks, surpassing the strongest competitor by $0.05$--$8.17$ percentage points, as shown in Table~\ref{tab:ovcd_seven_dataset_comparison}. It obtains $32.48\%$ and $25.36\%$ mIoU on SECOND and SCSCD, improving the previous best results by $0.81$ and $2.14$ points. For binary CD, CogVis reaches $33.67\%$, $48.50\%$, $72.55\%$, and $75.86\%$ IoU on CLCD, DSIFN, LEVIR-CD, and WHU-CD, corresponding to gains of $2.19$, $0.95$, $0.05$, and $0.14$ points. The larger gains on CLCD and DSIFN are consistent with the framework's ability to suppress appearance-induced temporal nuisances. The smaller margins on the cleaner LEVIR-CD and WHU-CD datasets reflect the already strong performance of existing methods. On xBD, CogVis improves mIoU from $24.51\%$ to $32.68\%$, producing the largest gain of $8.17$ points. These consistent improvements across all three tasks demonstrate the transferability of the perception-memory-verification design. \par

\textbf{\textit{{Qualitative Comparison:}}} As shown in Fig.~\ref{fig_3}, CogVis produces more coherent change regions and suppresses appearance-induced false responses across the seven benchmarks. It better separates queried categories on semantic and damage-assessment benchmarks while improving change localization on binary benchmarks. Figure~\ref{fig_4} further presents representative error maps under shadows, radiometric or textural variation, seasonal shifts, and residual misregistration. CogVis often yields sharper boundaries and predictions that more closely align with the annotations. These examples illustrate its robustness to non-semantic perturbations through change-prior modeling, query-specific calibration, and region-level reliability filtering.

\textbf{\textit{{Efficiency and Scalability:}}} CogVis reaches $0.230$ FPS with 4.82~GiB peak memory, providing $28.50\%$ higher throughput than the next-fastest method in Fig.~\ref{fig_5}. Since CogVis is not the smallest model, this gain arises from shared temporal computation rather than parameter reduction. At $K=10$, CogVis requires 10.03~s, compared with 12.56--956.60~s for category-wise baselines. Latency still increases with the number of queries because SMC and ARF remain query-specific, but one-time temporal perception substantially reduces the per-query cost.

\subsection{Ablation Studies}

\begin{figure}[t]
    \centering
    \includegraphics[width=\linewidth]{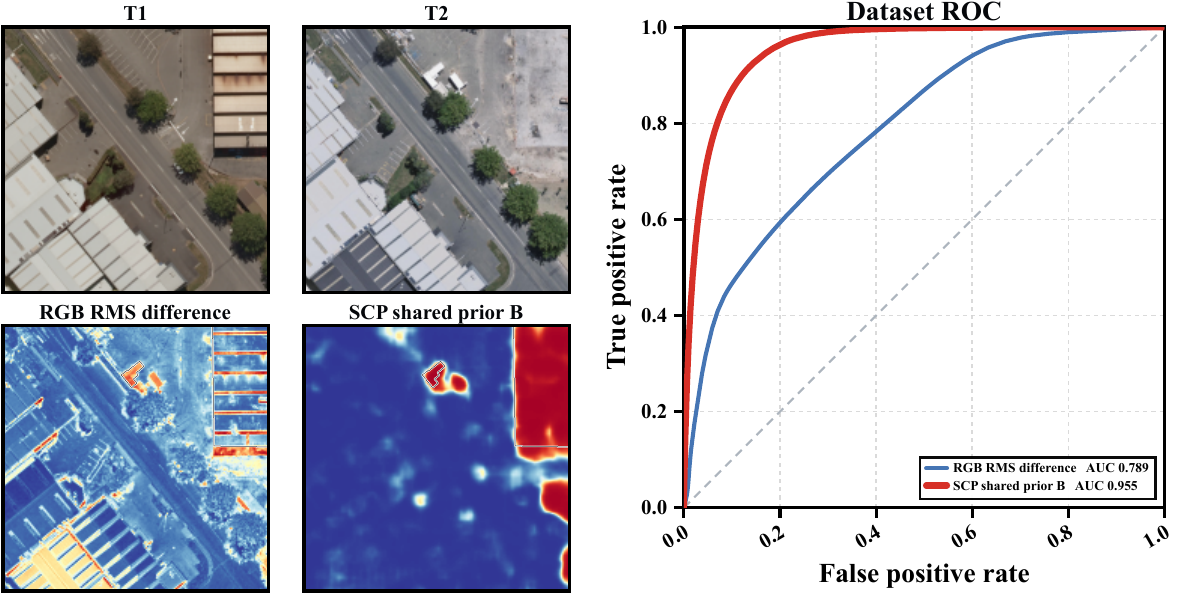}
    \caption{SCP analysis. The learned change prior suppresses appearance-only differences and improves ROC AUC from $0.79$ to $0.96$.}
    \label{fig_6}
\end{figure}

\textbf{\textit{{Component Ablation:}}} The full model outperforms Vanilla by $13.12$ mIoU on SECOND, $21.57$ IoU on CLCD, and $4.25$ mIoU on xBD, as shown in Table~\ref{tab:cogvis_ablation}. Among individual components, SCP delivers the largest gains on SECOND and CLCD with improvements of $2.95$ and $8.13$ points, whereas SMC contributes the largest gain on xBD at $1.85$ points. These results align with their respective roles in temporal stabilization and category-dependent calibration. Combining SCP and SMC produces the strongest two-component variant on all three benchmarks, confirming their complementarity. However, ARF depends strongly on the quality of upstream predictions. Used alone, it slightly reduces performance on SECOND while improving CLCD and xBD by $3.22$ and $0.17$ points. Combined only with SMC, it slightly degrades performance on SECOND and xBD but improves CLCD by $3.25$ points, indicating that region verification remains unstable without a reliable temporal prior. With both SCP and SMC enabled, ARF consistently improves SECOND, CLCD, and xBD by $4.44$, $4.76$, and $1.82$ points, respectively. These findings establish ARF as a final verification stage for temporally stabilized and semantically calibrated candidates rather than a generic post-processing module. \par

\begin{table}[t]
\centering
\small
\setlength{\tabcolsep}{1mm}
\begin{tabular}{@{}cccc!{\vrule width 0.30pt\hskip0.65pt\vrule width 0.30pt}cccc@{}}
\toprule[0.85pt]
\multirow[c]{2}{*}{\textbf{Variant}} &
\multicolumn{3}{c}{\textbf{Modules}} &
\multicolumn{1}{c}{\textbf{SECOND}} &
\multicolumn{1}{c}{\textbf{CLCD}} &
\multicolumn{1}{c}{\textbf{xBD}} &
\multirow[c]{2}{*}{\textbf{\#P $\downarrow$}} \\
\cmidrule(lr){2-4}\cmidrule(lr){5-5}\cmidrule(lr){6-6}\cmidrule(lr){7-7}
 & SCP & SMC & ARF & \textbf{mIoU $\uparrow$} & \textbf{IoU $\uparrow$} & \textbf{mIoU $\uparrow$} & \\
\midrule[0.60pt]\addlinespace[0.08em]
Vanilla &  &  &  & 19.36 & 12.10 & 28.43 & 840.51 \\
I  & \ensuremath{\checkmark} &  &  & 22.31 & 20.23 & 28.93 & 840.66 \\
II &  & \ensuremath{\checkmark} &  & 20.25 & 13.96 & 30.28 & 990.14 \\
III &  &  & \ensuremath{\checkmark} & 19.30 & 15.32 & 28.60 & 840.52 \\
\midrule[0.35pt]\addlinespace[0.12em]
IV & \ensuremath{\checkmark} & \ensuremath{\checkmark} &  & 28.04 & 28.91 & 30.86 & 990.29 \\
V & \ensuremath{\checkmark} &  & \ensuremath{\checkmark} & 25.75 & 25.87 & 29.12 & 840.66 \\
VI &  & \ensuremath{\checkmark} & \ensuremath{\checkmark} & 20.17 & 17.21 & 29.54 & 990.14 \\
\midrule[0.50pt]\addlinespace[0.10em]
\rowcolor[gray]{0.94}
\textbf{VII} & \ensuremath{\checkmark} & \ensuremath{\checkmark} & \ensuremath{\checkmark} & \textbf{32.48} & \textbf{33.67} & \textbf{32.68} & 990.29 \\
\bottomrule[0.85pt]
\end{tabular}
\normalsize
\caption{Component ablation. Bold marks the best result in each metric column. \#P denotes total parameters (M).}
\label{tab:cogvis_ablation}
\end{table}

\begin{figure}[t]
    \centering
    \includegraphics[width=\linewidth]{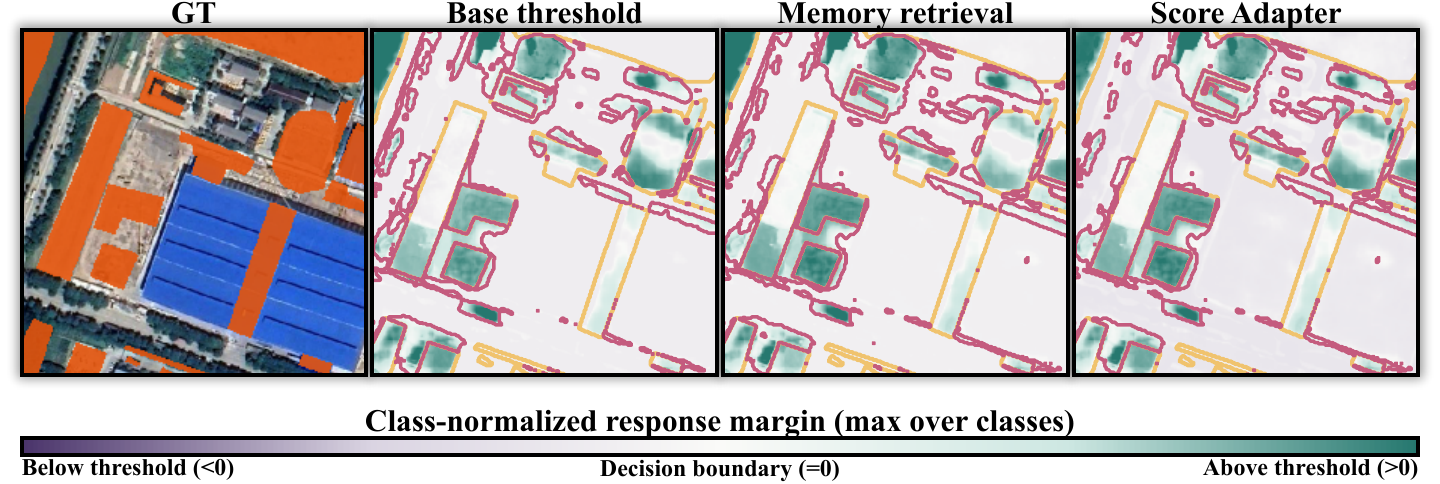}
    \caption{SMC analysis. Memory retrieval anchors the threshold, which the Score Adapter corrects for each image-query pair. Gold and magenta contours denote ground truth and false positives.}
    \label{fig_7}
\end{figure}

\begin{figure}[t]
    \centering
    \includegraphics[width=\linewidth]{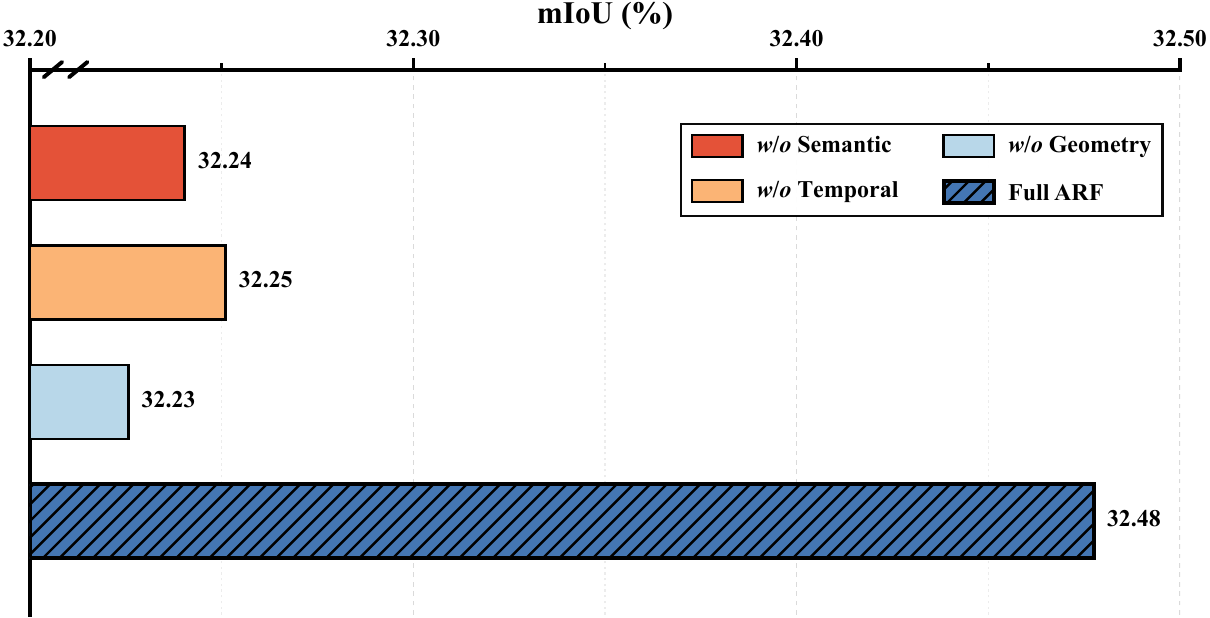}
    \caption{ARF evidence ablation on SECOND. Removing semantic, temporal, or geometric evidence degrades region verification.}
    \label{fig_8}
\end{figure}

\textbf{\textit{{Modular Ablation:}}} Figure~\ref{fig_6} shows that SCP increases ROC AUC from $0.79$ with raw RGB differences to $0.96$ with the learned change prior. Figure~\ref{fig_7} visualizes this progression: retrieval provides a suitable threshold anchor, and sample-specific correction suppresses residual false positives. Figure~\ref{fig_8} further shows consistent degradation when semantic, temporal, or geometric evidence is removed from ARF. Together, these results support the distinct and complementary roles of temporal perception, semantic calibration, and region verification. \par

\textbf{\textit{{Memory-Bank Ablation:}}} Table~\ref{tab:second_source_exclusion} tests whether CogVis's SECOND gains depend on the CA-CDD subset. Memory-only exclusion reduces SECOND/CLCD/xBD by $0.38$/$0.35$/$0.02$ points, while excluding the subset from adapter optimization and memory construction reduces them by $1.06$/$0.50$/$0.22$. These small, cross-benchmark decreases are more consistent with reduced source diversity and training volume than with a SECOND-specific advantage. This interpretation matches the adapter objectives, which target category-agnostic temporal evidence, residual threshold correction, and region reliability rather than dataset-specific semantics. Disabling the remaining cross-source memory causes further losses of $8.03$, $7.47$, and $1.60$ points, indicating that the gains derive from transferable response-threshold knowledge rather than same-dataset memorization. Table~\ref{tab:cogvis_memory_bank_scale_multidataset} shows that a $10\%$ bank retains $94.9\%$/$97.7\%$/$95.6\%$ of full-memory gains on SECOND/CLCD/xBD while reducing storage from $525.2$ to $52.6$ MiB. At $25\%$ or $50\%$, changes are marginal, and the small frozen/refitted gap indicates that saturation is not driven by scale-specific retraining, although the full bank remains best overall. Thus, SMC requires representative coverage of the response-threshold space rather than exhaustive source storage, making semantic memory a compact, transferable calibration prior.

\begin{table}[t]
\centering
\providecommand{\CogVisSecondCell}[1]{\underline{#1}}
\providecommand{\CogVisKeep}{\CogVisAblMark}
\providecommand{\CogVisDrop}{\ensuremath{\times}}
\providecommand{\CogVisAblDash}{--}
\small
\setlength{\tabcolsep}{1mm}
\begin{tabular}{@{}ccc!{\vrule width 0.30pt\hskip0.65pt\vrule width 0.30pt}ccc@{}}
\toprule[0.85pt]
\multirow[c]{2}{*}{\textbf{Variant}} &
\multicolumn{2}{c}{\textbf{CA-CDD}} &
\multicolumn{1}{c}{\textbf{SECOND}} &
\multicolumn{1}{c}{\textbf{CLCD}} &
\multicolumn{1}{c}{\textbf{xBD}} \\
\cmidrule(lr){2-3}\cmidrule(lr){4-4}\cmidrule(lr){5-5}\cmidrule(lr){6-6}
& Adapters & Memory & \textbf{mIoU $\uparrow$} & \textbf{IoU $\uparrow$} & \textbf{mIoU $\uparrow$} \\
\midrule[0.60pt]\addlinespace[0.08em]
I   & \CogVisKeep & \CogVisDrop & 32.10 & 33.32 & 32.66 \\
II  & \CogVisDrop & \CogVisDrop & 31.42 & 33.17 & 32.46 \\
III & \CogVisDrop & \CogVisAblDash & 23.39 & 25.70 & 30.86 \\
\midrule[0.50pt]\addlinespace[0.10em]
\rowcolor[gray]{0.94}
\textbf{Full} & \CogVisKeep & \CogVisKeep & \textbf{32.48} & \textbf{33.67} & \textbf{32.68} \\
\bottomrule[0.85pt]
\end{tabular}
\normalsize
\caption{Source-domain dependence. ``\CogVisKeep'' and ``\CogVisDrop'' indicate whether the subset of CA-CDD is retained or excluded from adapter training or memory construction. ``\CogVisAblDash'' disables memory entirely.
Bold marks the best result in each metric column.}
\label{tab:second_source_exclusion}
\end{table}

\begin{table}[t]
\centering
\small
\setlength{\tabcolsep}{1mm}
\begin{tabular}{@{}cc!{\vrule width 0.30pt\hskip0.65pt\vrule width 0.30pt}ccccc@{}}
\toprule[0.85pt]
\multirow[c]{2}{*}{\shortstack{\textbf{Memory}\\\textbf{bank}}} &
\multicolumn{1}{c@{}}{\multirow[c]{2}{*}{\textbf{Cal.}}} &
\multicolumn{1}{c}{\textbf{SECOND}} &
\multicolumn{1}{c}{\textbf{CLCD}} &
\multicolumn{1}{c}{\textbf{xBD}} &
\multicolumn{2}{c}{\textbf{Efficiency}} \\
\cmidrule(lr){3-3}\cmidrule(lr){4-4}\cmidrule(lr){5-5}\cmidrule(lr){6-7}
& & \textbf{mIoU $\uparrow$} & \textbf{IoU $\uparrow$} & \textbf{mIoU $\uparrow$} & \textbf{Mem. $\downarrow$} & \textbf{Lat. $\downarrow$} \\
\midrule[0.60pt]\addlinespace[0.08em]
\multirow[c]{2}{*}{0\%} & Fix. & 26.35 & 25.88 & 31.08 & \multirow[c]{2}{*}{0.0} & \multirow[c]{2}{*}{0} \\
& Rft. & -- & -- & -- & & \\
\cmidrule(lr){1-2}\cmidrule(lr){3-7}\addlinespace[0.08em]
\multirow[c]{2}{*}{10\%} & Fix. & 32.17 & 33.49 & 32.61 & \multirow[c]{2}{*}{52.6} & \multirow[c]{2}{*}{0.0012} \\
& Rft. & 32.00 & 33.18 & 32.64 & & \\
\cmidrule(lr){1-2}\cmidrule(lr){3-7}\addlinespace[0.08em]
\multirow[c]{2}{*}{25\%} & Fix. & 32.21 & 33.18 & 32.65 & \multirow[c]{2}{*}{131.4} & \multirow[c]{2}{*}{0.0030} \\
& Rft. & 32.11 & 32.91 & 32.66 & & \\
\cmidrule(lr){1-2}\cmidrule(lr){3-7}\addlinespace[0.08em]
\multirow[c]{2}{*}{50\%} & Fix. & 32.27 & 33.26 & 32.67 & \multirow[c]{2}{*}{262.6} & \multirow[c]{2}{*}{0.0097} \\
& Rft. & 32.29 & 32.97 & 32.66 & & \\
\midrule[0.50pt]\addlinespace[0.08em]
\rowcolor[gray]{0.94}
& Fix. & \textbf{32.48} & \textbf{33.67} & \textbf{32.68} & & \\
\rowcolor[gray]{0.94}
\raisebox{0.55em}[0pt][0pt]{\textbf{100\%}} & Rft. & \textbf{32.48} & \textbf{33.67} & \textbf{32.68} & \raisebox{0.55em}[0pt][0pt]{525.2} & \raisebox{0.55em}[0pt][0pt]{0.0113} \\
\bottomrule[0.85pt]
\end{tabular}
\normalsize
\caption{Memory-bank scale analysis. ``Cal." denotes calibration; ``Fix. and Rft." denote frozen and refitted calibration, respectively, while Mem. (MiB) and Lat. (ms/query) report storage and retrieval latency. Bold marks the best accuracy in each metric column.}
\label{tab:cogvis_memory_bank_scale_multidataset}
\end{table}

\subsection{Sensitivity Analysis}
Table~\ref{tab:cogvis_hyperparameter_sensitivity} shows that CogVis is robust to memory-retrieval settings: varying $K_{\mathrm{mem}}$ from 8 to 32 or $T_m$ from 0.05 to 0.10 causes only marginal changes across all benchmarks. Consistent with the memory-bank ablation, this stability indicates that SMC transfers representative response-threshold relationships without relying on a specific retrieval neighborhood. Region-level parameters have larger effects, particularly on CLCD: an overly permissive $\tau_r$ retains unreliable components, whereas an overly strict $\tau_r$ removes valid changes. Lowering $\eta_a$ also makes large-region protection overly permissive, allowing smaller unreliable components to bypass the learned gate and reducing performance on SECOND and CLCD. The default configuration offers the best overall balance, supporting CogVis's intended functional separation.

\begin{table}[!b]
\centering
\small
\setlength{\tabcolsep}{1mm}
\begin{tabular}{@{}ccccc!{\vrule width 0.30pt\hskip0.65pt\vrule width 0.30pt}ccc@{}}
\toprule[0.85pt]
\multirow[c]{2}{*}{\textbf{Variant}} &
\multicolumn{4}{c}{\textbf{Configuration}} &
\multicolumn{1}{c}{\textbf{SECOND}} &
\multicolumn{1}{c}{\textbf{CLCD}} &
\multicolumn{1}{c}{\textbf{xBD}} \\
\cmidrule(lr){2-5}\cmidrule(lr){6-6}\cmidrule(lr){7-7}\cmidrule(lr){8-8}
& $K_{\mathrm{mem}}$ & $T_m$ & $\tau_r$ & $\eta_a$ & \textbf{mIoU $\uparrow$} & \textbf{IoU $\uparrow$} & \textbf{mIoU $\uparrow$} \\
\midrule[0.60pt]\addlinespace[0.08em]
I  & 8  & 0.07 & 0.6 & 0.04 & 32.30 & 33.17 & 32.67 \\
II & 32 & 0.07 & 0.6 & 0.04 & 32.32 & 33.45 & 32.68 \\
\midrule[0.35pt]\addlinespace[0.12em]
III & 16 & 0.05 & 0.6 & 0.04 & 32.29 & 33.32 & 32.67 \\
IV  & 16 & 0.10 & 0.6 & 0.04 & 32.29 & 33.33 & 32.67 \\
\midrule[0.35pt]\addlinespace[0.12em]
V  & 16 & 0.07 & 0.5 & 0.04 & 32.16 & 32.05 & 32.32 \\
VI & 16 & 0.07 & 0.7 & 0.04 & 32.17 & 32.33 & 32.62 \\
\midrule[0.35pt]\addlinespace[0.12em]
VII  & 16 & 0.07 & 0.6 & 0.01 & 32.11 & 32.69 & 32.58 \\
\rowcolor[gray]{0.94}
\textbf{Default} & 16 & 0.07 & 0.6 & 0.04 & \textbf{32.48} & \textbf{33.67} & \textbf{32.68} \\
\bottomrule[0.85pt]
\end{tabular}
\normalsize
\caption{Hyperparameter sensitivity. Bold marks the best accuracy in each metric column, and the bold row denotes the default configuration.}
\label{tab:cogvis_hyperparameter_sensitivity}
\end{table}

\section{Conclusion}
This work revisits OVCD from the perspective of decision scope. CogVis separates query-shared temporal perception from query-dependent semantic calibration and region-level verification, allowing each source of uncertainty to be addressed at the appropriate stage. Through SCP, SMC, and ARF, it reuses scene-level change evidence, transfers response-threshold knowledge across domains, and suppresses unreliable regions using complementary semantic, temporal, and structural cues. Experiments on seven benchmarks demonstrate consistent state-of-the-art performance, strong cross-task transferability, and a $28.50\%$ improvement in inference throughput. These results show that open-vocabulary change understanding need not reinterpret the scene for every query.

\makeatletter
\let\CogVisOriginalLbibitem\@lbibitem
\def\@lbibitem[#1]#2{%
  \def\CogVisCurrentBibKey{#2}%
  \def\CogVisBalanceAtKey{wang2025accurate}%
  \ifx\CogVisCurrentBibKey\CogVisBalanceAtKey
    \newpage
  \fi
  \CogVisOriginalLbibitem[#1]{#2}%
}
\makeatother
\bibliography{CogVisReferences}

\appendix
\setcounter{secnumdepth}{2}
\renewcommand{\thesection}{\Alph{section}}
\renewcommand{\thesubsection}{\thesection.\arabic{subsection}}

\setcounter{equation}{0}
\setcounter{figure}{0}
\setcounter{table}{0}
\renewcommand{\theequation}{S\arabic{equation}}
\renewcommand{\thefigure}{S\arabic{figure}}
\renewcommand{\thetable}{S\arabic{table}}
\renewcommand{\CogVisTableGroupRule}{\midrule[0.35pt]\addlinespace[0.10em]}

\makeatletter
\setlength{\@fptop}{0pt}
\setlength{\@fpsep}{10pt plus 2pt minus 2pt}
\setlength{\@fpbot}{0pt plus 1fil}
\setlength{\@dblfptop}{0pt}
\setlength{\@dblfpsep}{10pt plus 2pt minus 2pt}
\setlength{\@dblfpbot}{0pt plus 1fil}
\makeatother
\renewcommand{\topfraction}{0.95}
\renewcommand{\bottomfraction}{0.90}
\renewcommand{\textfraction}{0.05}
\renewcommand{\floatpagefraction}{0.80}
\renewcommand{\dbltopfraction}{0.95}
\renewcommand{\dblfloatpagefraction}{0.72}
\setcounter{topnumber}{4}
\setcounter{bottomnumber}{2}
\setcounter{totalnumber}{6}
\setcounter{dbltopnumber}{3}
\setlength{\textfloatsep}{11pt plus 2pt minus 2pt}
\setlength{\floatsep}{9pt plus 2pt minus 2pt}

\CogVisSupplementaryTitle

This supplement provides the implementation and evaluation details omitted
from the main paper. Section~\ref{sec:supp_experimental_details} specifies the
benchmark interfaces and prompt vocabularies.
Section~\ref{sec:supp_architecture} expands the compact main-paper equations
into an executable, fully parameterized description.
Section~\ref{sec:supp_detailed_results} reports class-resolved results together
with additional query- and region-level analyses, and
Section~\ref{sec:supp_wildfire_case} examines the same fixed inference rule in
kilometer-scale scenes.

\section{Experimental Protocol}
\label{sec:supp_experimental_details}

All experiments used $512\times512$ inputs and a dataset-specific prompt
vocabulary that remained fixed across the complete test split. The adapters
were trained and semantic memory was constructed using only CA-CDD; no
target-benchmark test image contributed to optimization or memory construction.
The resulting protocol evaluates cross-benchmark transfer under fixed
text-specified vocabularies.

\subsection{Benchmarks and Evaluation Protocol}

The seven benchmarks span semantic, binary, and damage-state change detection.
SECOND \citep{yang2022second} and SCSCD \citep{tan2025triples} require one
semantic-change mask per foreground query; SCSCD is evaluated with its seven
foreground classes. CLCD \citep{liu2022clcd} merges the
queried land-cover masks into a binary map. DSIFN \citep{zhang2020dsifn},
LEVIR-CD \citep{chen2020levircd}, and WHU-CD \citep{ji2019whu} instead treat
buildings as foreground. WHU-CD is the bitemporal change subset of the WHU
Building Dataset. xBD \citep{gupta2019xbd} evaluates building support and four
post-event damage states. SECOND and SCSCD use macro-averaged metrics over
foreground classes, binary benchmarks use foreground IoU and F1, and xBD uses
five-state macro metrics that include background.

\subsection{Evaluation Metrics}
\label{sec:supp_metrics}

Let $\mathcal{Y}_c$ and $\widehat{\mathcal{Y}}_c$ denote the reference and
predicted pixel sets for class $c$. The corresponding confusion terms are
\begin{equation}
\begin{aligned}
\mathrm{TP}_c
&=|\mathcal{Y}_c\cap\widehat{\mathcal{Y}}_c|,\\
\mathrm{FP}_c
&=|\widehat{\mathcal{Y}}_c\setminus\mathcal{Y}_c|,
\qquad
\mathrm{FN}_c=|\mathcal{Y}_c\setminus\widehat{\mathcal{Y}}_c|.
\end{aligned}
\label{eq:supp_confusion_counts}
\end{equation}
Class-wise and macro-averaged scores are
\begin{equation}
\begin{aligned}
\operatorname{IoU}_c
&=\frac{\mathrm{TP}_c}
{\mathrm{TP}_c+\mathrm{FP}_c+\mathrm{FN}_c},\\
\operatorname{F1}_c
&=\frac{2\mathrm{TP}_c}
{2\mathrm{TP}_c+\mathrm{FP}_c+\mathrm{FN}_c},\\
\operatorname{mIoU}
&=\frac{1}{|\mathcal{C}_{\rm eval}|}
\sum_{c\in\mathcal{C}_{\rm eval}}\operatorname{IoU}_c,\\
\operatorname{mF1}
&=\frac{1}{|\mathcal{C}_{\rm eval}|}
\sum_{c\in\mathcal{C}_{\rm eval}}\operatorname{F1}_c.
\end{aligned}
\label{eq:supp_accuracy_metrics}
\end{equation}
For xBD, predictions are evaluated in the standard five-state output space
comprising background, no damage, minor damage, major damage, and destroyed.
The reported mIoU and mF1 are macro-averaged over these five states after final
background assignment. To resolve performance by damage category, we also
report the F1 score of each damage state on reference building pixels before
background assignment. The same class-wise protocol is applied to every method
in the comparison.

Efficiency was measured after model initialization with a batch size of one.
For $N$ image pairs processed in time $T$, latency and throughput are
\begin{equation}
\begin{aligned}
\operatorname{Latency}
&=\frac{T}{N},\\
\operatorname{FPS}
&=\frac{N}{T}.
\end{aligned}
\label{eq:supp_efficiency_metrics}
\end{equation}
Peak memory denotes the maximum GPU allocation during inference, and parameter
counts are reported in millions. Query-scaling measurements used one warm-up
and one synchronized full-request measurement for each vocabulary size, with
model loading and image-disk input excluded. The profiling environment used
Python~3.10.20, PyTorch~2.5.1 with CUDA~12.1 and an RTX~3090 GPU; inference was
measured in FP32 with TF32 disabled.

\paragraph{Evaluation scope.}
All results are evaluated over the complete test splits using fixed checkpoints
and benchmark-stationary prompt vocabularies. Class-wise outcomes are
interpreted jointly through absolute margins and cross-category consistency.

\paragraph{Multi-query computational accounting.}
CogVis separates computation shared by an image pair from computation repeated
for each query. For a vocabulary of $K$ queries, latency is
\begin{equation}
T(K)=T_{\rm pair}+K\,T_{\rm query},
\label{eq:supp_query_cost}
\end{equation}
where $T_{\rm pair}$ comprises frozen feature extraction, SCP, and pair
embeddings, while $T_{\rm query}$ comprises semantic decoding, memory
retrieval, and region verification. Equation~\eqref{eq:supp_query_cost}
records the accounting used for the main-paper efficiency study without
repeating its measured latency, throughput, and memory results.

\subsection{Text Prompt Vocabulary}
\label{sec:supp_prompt_vocabulary}

Each prompt vocabulary contains the canonical class name, common
remote-sensing synonyms, and variants that describe distinct visual forms.
The frozen prompt decoder encodes each phrase independently, and phrase
responses within a class are combined by a pixelwise maximum. The vocabulary
remains unchanged for every image in a benchmark, with no image-specific prompt
selection. In Tables~\ref{tab:supp_semantic_prompts}--\ref{tab:supp_damage_prompts},
braces group phrases assigned to one class, and bold labels denote evaluated
foreground queries.

\begin{table}[!t]
\centering
\CogVisPromptTableSetup
\begin{tabular}{@{}L{0.17\columnwidth}L{0.24\columnwidth}!{\CogVisDoubleSep}L{0.50\columnwidth}@{}}
\CogVisTableTopRule
\textbf{Dataset} & \textbf{Query class} & \textbf{Text prompts} \\
\CogVisTableHeaderRule
\multirow{7}{*}{SECOND}
& Background & \{background\} \\
& \textbf{Water} & \{water body, river, lake, reservoir\} \\
& \textbf{Surface} & \{bare land, bare soil, impervious barren ground, exposed ground\} \\
& \textbf{Low vegetation} & \{low vegetation, grass, lawn, shrub, grassland, meadow\} \\
& \textbf{Tree} & \{tree canopy, forest, woodland, trees\} \\
& \textbf{Building} & \{building, roof, house, structure\} \\
& \textbf{Playground} & \{playground, sports field, running track, athletic field\} \\
\CogVisTableGroupRule
\multirow{7}{*}{SCSCD}
& \textbf{Bareland} & \{bare land, bare soil, exposed ground, barren land\} \\
& \textbf{Water} & \{water body, river, pond, canal\} \\
& \textbf{Building} & \{building, rooftop, house, urban structure\} \\
& \textbf{Structure} & \{greenhouse, construction structure, built structure\} \\
& \textbf{Farmland} & \{farmland, cropland, agricultural field\} \\
& \textbf{Vegetation} & \{vegetation, tree canopy, grassland, forest\} \\
& \textbf{Road} & \{road, paved road, transportation road\} \\
\CogVisTableBottomRule
\end{tabular}
\normalsize
\caption{Prompt vocabulary for semantic change detection. Braces group phrases whose responses are aggregated into one class map.}
\label{tab:supp_semantic_prompts}
\end{table}

\begin{table}[!t]
\centering
\CogVisPromptTableSetup
\begin{tabular}{@{}L{0.18\columnwidth}L{0.24\columnwidth}!{\CogVisDoubleSep}L{0.48\columnwidth}@{}}
\CogVisTableTopRule
\textbf{Dataset} & \textbf{Query class} & \textbf{Text prompts} \\
\CogVisTableHeaderRule
\multirow{7}{*}{CLCD}
& \textbf{Bareland} & \{bare land, bare soil, exposed ground\} \\
& \textbf{Grass} & \{low vegetation, grassland\} \\
& \textbf{Road} & \{road, paved road\} \\
& \textbf{Tree} & \{tree canopy, forest\} \\
& \textbf{Water} & \{water body, river, lake\} \\
& \textbf{Cropland} & \{cropland, farmland, agricultural field\} \\
& \textbf{Building} & \{building, rooftop, urban structure\} \\
\CogVisTableGroupRule
\multirow{2}{*}{DSIFN}
& Background & \{ground, road, vegetation, farmland, bare ground, water\} \\
& \textbf{Building} & \{urban building, building, building rooftop, roof, built-up area, urban structure\} \\
\CogVisTableGroupRule
\multirow{2}{*}{LEVIR-CD}
& Background & \{ground, road, pavement, impervious surface, vegetation, water\} \\
& \textbf{Building} & \{building, roof\} \\
\CogVisTableGroupRule
\multirow{2}{*}{WHU-CD}
& Background & \{ground, road, vegetation, asphalt, pavement\} \\
& \textbf{Building} & \{building, roof\} \\
\CogVisTableBottomRule
\end{tabular}
\normalsize
\caption{Prompt vocabulary for binary change detection. CLCD merges seven class maps into a binary map; DSIFN, LEVIR-CD and WHU-CD use Building as the foreground query.}
\label{tab:supp_binary_prompts}
\end{table}

\begin{table}[!t]
\centering
\CogVisPromptTableSetup
\begin{tabular}{@{}L{0.18\columnwidth}L{0.25\columnwidth}!{\CogVisDoubleSep}L{0.48\columnwidth}@{}}
\CogVisTableTopRule
\textbf{Dataset} & \textbf{Query class} & \textbf{Text prompts} \\
\CogVisTableHeaderRule
\multirow{5}{*}{xBD}
& \textbf{Building support} & \{building, roof, house, building footprint, standing building\} \\
& \textbf{No damage} & \{undamaged building, intact building, standing intact building\} \\
& \textbf{Minor damage} & \{slightly damaged building, minor roof damage, limited visible damage\} \\
& \textbf{Major damage} & \{heavily damaged building, severely damaged building, partially collapsed building, major structural damage\} \\
& \textbf{Destroyed} & \{collapsed building, destroyed building, completely collapsed building, building rubble, debris of a building\} \\
\CogVisTableBottomRule
\end{tabular}
\normalsize
\caption{Prompt vocabulary for building localization and damage assessment on xBD.}
\label{tab:supp_damage_prompts}
\end{table}

\section{Model Specification}
\label{sec:supp_architecture}

CogVis aligns each learned component with a distinct decision scale. The Scene
Change Perceptron (SCP) extracts temporal evidence once per image pair. The
Semantic Memory Calibrator (SMC) converts frozen semantic responses into a
query-specific response and threshold. The Adaptive Region Filter (ARF)
evaluates connected candidates at region level. We next define the full
computation and the objectives for the Change, Score, and Gate Adapters.

\subsection{End-to-End Computation}
\label{sec:supp_computation_flow}

For registered images $I^1,I^2\in\mathbb{R}^{H\times W\times3}$, the frozen
visual encoder \citep{carion2025sam3} produces
\begin{equation}
\begin{aligned}
\{\mathbf{F}_l^t\}_{l=1}^{L}
&=\mathcal{E}(I^t),\qquad t\in\{1,2\},\\
\mathcal{S}_{12}
&=\bigl(\{\mathbf{F}_l^1,\mathbf{F}_l^2\}_{l=1}^{L},
\mathbf{B},\mathbf{e}_s,\mathbf{e}_d\bigr).
\end{aligned}
\label{eq:supp_pair_state}
\end{equation}
Here, $\mathbf{B}$ is the SCP prior, and $\mathbf{e}_s$ and $\mathbf{e}_d$ are
scene and temporal CLIP embeddings \citep{radford2021clip}. The resulting pair
state, $\mathcal{S}_{12}$, is computed once and reused for all queries.

For phrase $p$ at time $t$, the frozen prompt decoder returns a semantic
probability map $\mathbf{s}_p^t$ and a presence score $\varpi_p^t$. When
activated, the instance branch also returns masks $\mathbf{m}_{p,k}^t$ and
confidences $o_{p,k}^t$. The phrase-level, class-level target, and strongest
competing responses are
\begin{equation}
\begin{aligned}
\mathbf{q}_p^t
&=\max\!\left(\varpi_p^t\mathbf{s}_p^t,
\max_k o_{p,k}^t\mathbf{m}_{p,k}^t\right),\\
\mathbf{Q}_c^t
&=\max_{p\in\mathcal{P}_c}\mathbf{q}_p^t,\\
\mathbf{C}_c^t
&=\max_{n\in\mathcal{N}_c}
\max_{p\in\mathcal{P}_n}\mathbf{q}_p^t.
\end{aligned}
\label{eq:supp_semantic_inputs}
\end{equation}
The corresponding scalar class-presence score is
$\pi_c^t=\max_{p\in\mathcal{P}_c}\varpi_p^t$.
SMC maps the endpoint responses and $\mathbf{B}$ to a continuous change
response $\mathbf{R}_c$ and threshold $\tau_c$. ARF filters the connected
components of
$\widetilde{\mathbf{M}}_c=\mathbf{1}[\mathbf{R}_c\ge\tau_c]$:
\begin{equation}
(I^1,I^2)\xrightarrow{\rm SCP}\mathbf{B}
\xrightarrow[\,c\,]{\rm SMC}(\mathbf{R}_c,\tau_c)
\xrightarrow{\rm ARF}\mathbf{M}_c.
\label{eq:supp_cogvis_flow}
\end{equation}

\subsection{Change Adapter and Scene Change Perceptron}
\label{sec:supp_change_adapter}

\textbf{Construction.}
Here $\operatorname{Cat}_{\rm ch}(\cdot)$ denotes channel-wise concatenation
along the feature dimension, and
$\mathcal{H}_{\rm chg}(\mathbf{X},\boldsymbol{\Delta})$ denotes the change
head, a shallow convolutional network (see Table~\ref{tab:supp_adapter_architecture})
that maps the paired descriptor and temporal residual to a spatial change
logit map.
SCP operates on the second level of the frozen feature pyramid,
$\mathbf{F}^{t}=\mathbf{F}_{2}^{t}$, and constructs a paired descriptor from
endpoint appearance, absolute discrepancy and feature agreement:
\begin{equation}
\begin{aligned}
\mathbf{X}
&=\operatorname{Cat}_{\rm ch}\!\left(
\mathbf{F}^{1},\mathbf{F}^{2},
|\mathbf{F}^{2}-\mathbf{F}^{1}|,
\mathbf{F}^{1}\odot\mathbf{F}^{2}\right),\\
\boldsymbol{\Delta}
&=\mathbf{W}_{\uparrow}\phi(\mathbf{W}_{\downarrow}\mathbf{X}).
\end{aligned}
\label{eq:supp_change_adapter}
\end{equation}
The low-rank Change Adapter produces the temporal residual
$\boldsymbol{\Delta}$. A temporal gate limits its root-mean-square magnitude to
$0.25$ of the frozen feature magnitude. The change head then predicts
\begin{equation}
\mathbf{B}
 =\sigma\!\left(
 \operatorname{Up}\bigl(
 \mathcal{H}_{\rm chg}(\mathbf{X},\boldsymbol{\Delta})
 \bigr)\right).
\label{eq:supp_scp_prior}
\end{equation}
The category-agnostic prior is computed once per image pair and reused for all
queries. The residual contributes only to $\mathbf{B}$; semantic decoding uses
the unmodified frozen feature pyramid.

\textbf{Optimization.}
The Change Adapter is trained on binary source masks with equally weighted
class-balanced binary cross-entropy and soft Dice losses:
\begin{equation}
\mathcal{L}_{\rm chg}
=\mathcal{L}_{\rm bal\text{-}bce}(\mathbf{B},\mathbf{Y})
+\mathcal{L}_{\rm dice}(\mathbf{B},\mathbf{Y}).
\label{eq:supp_change_total_loss}
\end{equation}
The BCE positive-class weight is the mini-batch negative-to-positive pixel
ratio clipped to $[1,20]$. Source masks are resized to the change-logit
resolution by nearest-neighbor interpolation.

\subsection{Semantic Memory Calibrator and Score Adapter}
\label{sec:supp_score_adapter}

\textbf{Semantic-change response.}
Prompt competition downweights target activation that is also explained by
nearby semantic classes. The competition-adjusted endpoint responses define
absolute and directional semantic-transition evidence:
\begin{equation}
\begin{aligned}
\widetilde{\mathbf{Q}}_c^t
&=\mathbf{Q}_c^t\odot
\left(\frac{\mathbf{Q}_c^t}
{\mathbf{Q}_c^t+\mathbf{C}_c^t+\epsilon}\right)^\rho,\\
\mathbf{D}_c
&=\lambda_{\rm abs}\left|\widetilde{\mathbf{Q}}_c^2-
\widetilde{\mathbf{Q}}_c^1\right|
+\lambda_{\rm dir}\max_{t\ne t'}
\widetilde{\mathbf{Q}}_c^t\odot
(1-\widetilde{\mathbf{Q}}_c^{t'}),\\
\mathbf{R}_c
&=\clip_{[0,1]}\!\left(
\mathbf{D}_c\odot\mathbf{G}_c\odot\mathbf{S}_c\,p_c
\right).
\end{aligned}
\label{eq:supp_complete_response}
\end{equation}
The response modulation terms are
\begin{equation}
\begin{aligned}
\mathbf{G}_c
&=\epsilon_g+(1-\epsilon_g)\mathbf{B},\\
\mathbf{S}_c
&=\clip_{[0,1]}\!\left(
1-\lambda_s\min_t(\widetilde{\mathbf{Q}}_c^t)
\odot(1-\mathbf{B})\right),\\
p_c
&=\lvert\pi_c^2-\pi_c^1\rvert.
\end{aligned}
\label{eq:supp_response_modulation}
\end{equation}
The fixed coefficients are
$\epsilon=10^{-6}$, $\rho=2$,
$\lambda_{\rm abs}=0.55$, $\lambda_{\rm dir}=0.25$,
$\epsilon_g=0.10$, and $\lambda_s=0.45$. Thus, $\mathbf{R}_c$ retains
query-specific transitions only when semantic transition, shared temporal
support, stable-region suppression, and endpoint presence change agree.

\textbf{Memory retrieval.}
All spatial maps are bilinearly resized to the $512\times512$ inference grid.
For a map $\mathbf{X}$, let $Q_q(\mathbf{X})$ denote its spatial $q$-quantile
and let $\operatorname{Mass}_{>a}(\mathbf{X})$ be the fraction of pixels whose
value exceeds $a$. The 20-dimensional response statistic is
\begin{equation}
\begin{aligned}
\mathbf{r}_c={}&[\mu,\sigma,Q_{0.50},Q_{0.75},Q_{0.85},Q_{0.90},
Q_{0.95},Q_{0.98},\max,\\
&\operatorname{Mass}_{>\mu},
\operatorname{Mass}_{>Q_{0.85}}](\mathbf{R}_c),\\
\mathbf{b}={}&[\mu,\sigma,Q_{0.75},Q_{0.85},Q_{0.95},\max](\mathbf{B}),\\
\mathbf{u}_c={}&[\mu,Q_{0.85}](\mathbf{R}_c\odot\mathbf{B}),\\
\boldsymbol{\xi}_c={}&[\mathbf{r}_c,\mathbf{b},\mathbf{u}_c,
\pi_c^{\max}]\in\mathbb{R}^{20},
\end{aligned}
\label{eq:supp_response_statistics}
\end{equation}
where $\pi_c^{\max}=\max(\pi_c^1,\pi_c^2)$. CLIP ViT-B/16 provides normalized
endpoint embeddings $\mathbf{v}_1,\mathbf{v}_2\in\mathbb{R}^{512}$, from
which
\begin{equation}
\begin{aligned}
\mathbf{e}_s
&=\operatorname{Norm}\!\left(
\frac{\mathbf{v}_1+\mathbf{v}_2}{2}\right),\\
\mathbf{e}_d
&=\operatorname{Norm}(|\mathbf{v}_2-\mathbf{v}_1|).
\end{aligned}
\label{eq:supp_clip_pair_embeddings}
\end{equation}
The query embedding $\mathbf{e}_c$ is the normalized embedding of the first
registered phrase for class $c$. With
$\operatorname{Norm}(\mathbf{x})=\mathbf{x}/(\|\mathbf{x}\|_2+\epsilon)$,
the retrieval key is
\begin{equation}
\mathbf{k}_c=\operatorname{Norm}\!\left(
[\mathbf{e}_s,0.70\mathbf{e}_d,\mathbf{e}_c,
0.25\boldsymbol{\xi}_c]\right)\in\mathbb{R}^{1556}.
\label{eq:supp_memory_key}
\end{equation}
The memory enumerates 4,968 CA-CDD pairs and 12 registered semantic anchors,
producing 59,616 rows. The source operating point is selected over
\begin{equation}
\mathcal{G}_{\tau}=\{0.02,0.04,\ldots,0.80\}
\end{equation}
by
\begin{equation}
\tau_i^\star
=\arg\max_{\tau\in\mathcal{G}_{\tau}}
\operatorname{IoU}\!\left(
\mathbf{1}[\mathbf{R}_i\ge\tau],\mathbf{Y}_i\right).
\label{eq:supp_oracle_threshold}
\end{equation}
An empty source mask is assigned $\tau_i^\star=0.02$, and exact ties are
resolved by the smallest threshold. Cosine similarity retrieves the top 16
source-memory entries, denoted $\mathcal{K}_c$. After subtracting the largest
similarity for numerical stability, temperature weighting with $T_m=0.07$
gives
\begin{equation}
\begin{aligned}
\alpha_i^c
&=\frac{\exp((s_i-s_{\max})/T_m)}
{\sum_{j\in\mathcal{K}_c}\exp((s_j-s_{\max})/T_m)},\\
\bar{\tau}_c
&=\sum_{i\in\mathcal{K}_c}\alpha_i^c\tau_i^\star,
\end{aligned}
\label{eq:supp_memory_anchor}
\end{equation}
where $s_i=\mathbf{k}_c^\top\mathbf{k}_i$.

\textbf{Confidence-aware threshold correction.}
The label-free fallback $\mathcal{T}_{\rm adp}$ is deterministic. Let
$\mathcal{O}(\mathbf{R})$ denote 128-bin global Otsu thresholding. An
edge-aware value $\mathcal{O}_{\rm edge}(\mathbf{R})$ averages this threshold
with Otsu computed on a one-step $3\times3$ dilation of the top-20\%
Sobel-gradient support. It reverts to $\mathcal{O}(\mathbf{R})$ when the
maximum gradient is below $10^{-8}$ or the dilated support contains fewer than
64 pixels. Then
\begin{equation}
\begin{aligned}
\mathcal{T}_{\rm adp}(\mathbf{R})
&=\clip_{[0,1]}\bigl(\operatorname{median}\{
\mathcal{O}_{\rm edge}(\mathbf{R}),\\
&\qquad Q_{0.85}(\mathbf{R}),
\mu(\mathbf{R})+\sigma(\mathbf{R})\}\bigr).
\end{aligned}
\label{eq:supp_adaptive_threshold}
\end{equation}
The Score Adapter receives the 20 response statistics, memory anchor, top-1
similarity, and mean retrieved similarity as a 23-dimensional input:
\begin{equation}
\begin{aligned}
\mathbf{z}_c
&=[\boldsymbol{\xi}_c,\bar{\tau}_c,
 s_c^{\max},s_c^{\rm mean}],\\
\Delta\tau_c
&=0.20\tanh\mathcal{A}_{\rm score}(\mathbf{z}_c),\\
\mathcal{L}_{\rm score}
&=\frac{1}{N}\sum_i
\ell_{\rm sl1}\!\left(
\Delta\tau_i,\tau_i^\star-\bar{\tau}_i^{(-i)}\right),\\
\tau_c
&=\clip_{[0.02,0.80]}\!\left(
\max\left\{\bar{\tau}_c+\Delta\tau_c,
0.75\mathcal{T}_{\rm adp}(\mathbf{R}_c)\right\}\right).
\end{aligned}
\label{eq:supp_score_adapter}
\end{equation}
Leave-one-out retrieval removes only the current row and retains all remaining
rows, including the other semantic anchors from the same source pair. At
inference, the memory anchor transfers a source operating point, the bounded
correction adapts it to the current image--query pair, and the deterministic
scaled adaptive estimate $0.75\mathcal{T}_{\rm adp}(\mathbf{R}_c)$ supplies a conservative lower bound.

\begin{table*}[!t]
\centering
\CogVisCompactTableSetup
\begin{tabular}{@{}L{0.15\textwidth}L{0.15\textwidth}!{\CogVisDoubleSep}L{0.48\textwidth}C{0.12\textwidth}@{}}
\CogVisTableTopRule
\textbf{Module} & \textbf{Input} & \textbf{Ordered operators} & \textbf{Trainable parameters} \\
\CogVisTableHeaderRule
SCP adapter & $1024\times H_2\times W_2$ & Conv$_{1\times1}$ $1024\rightarrow16$, GELU, Conv$_{1\times1}$ $16\rightarrow256$; no bias; rank and scale 16 & 20,480 \\
SCP head and gain & Four spatial maps & Conv$_{3\times3}$ $4\rightarrow96$, GN$_8$, GELU, Conv$_{3\times3}$ $96\rightarrow96$, GN$_8$, GELU, Conv$_{1\times1}$ $96\rightarrow1$, scalar gain & 86,882 \\
Score Adapter & $\mathbf{z}_c\in\mathbb{R}^{23}$ & Linear$(23,64)$, LN, ReLU, Dropout$(0.05)$, Linear$(64,64)$, ReLU, Linear$(64,1)$ & 5,889 \\
Gate Adapter & $\boldsymbol{\varphi}_j\in\mathbb{R}^{19}$ & Linear$(19,64)$, LN, ReLU, Dropout$(0.05)$, Linear$(64,64)$, ReLU, Linear$(64,1)$, sigmoid & 5,633 \\
\CogVisTableBottomRule
\end{tabular}
\normalsize
\caption{Trainable architecture of SCP, SMC, and ARF. The counts cover the executed adapters and heads rather than the frozen foundation models.}
\label{tab:supp_adapter_architecture}
\end{table*}

\subsection{Adaptive Region Filter and Gate Adapter}
\label{sec:supp_gate_adapter}

\textbf{Region representation.}
Thresholding $\mathbf{R}_c$ yields
$\Omega_c=\operatorname{CCL}(\widetilde{\mathbf{M}}_c)$. For component
$\omega_j$, let $a_j=|\omega_j|$, $b_j=|\operatorname{bbox}(\omega_j)|$,
and $A=HW$. Here, $\operatorname{CCL}$ denotes 8-connected component
labeling, $\operatorname{bbox}$ the tight bounding box, and
$\operatorname{Top}_{0.20,j}$ the mean of the largest
$\max(1,\lfloor0.20a_j\rfloor)$ component values. Statistics restricted to
$\omega_j$ are summarized by
\begin{equation}
\begin{aligned}
\mathbf{r}_j(\mathbf{X})
&=[\mu_j(\mathbf{X}),\sigma_j(\mathbf{X}),
\max_j(\mathbf{X}),\operatorname{Top}_{0.20,j}(\mathbf{X})],\\
\operatorname{Mass}_j(\mathbf{X};\theta)
&=a_j^{-1}|\{x\in\omega_j:\mathbf{X}(x)>\theta\}|.
\end{aligned}
\label{eq:supp_region_statistics}
\end{equation}
With $\theta_B=\mathcal{T}_{\rm adp}(\mathbf{B})$, the four descriptor blocks
are
\begin{equation}
\begin{aligned}
\boldsymbol{\varphi}_j^{\rm geo}
&=[a_j/A,\sqrt{a_j/A},b_j/A,a_j/b_j],\\
\boldsymbol{\varphi}_j^{\rm sem}
&=\mathbf{r}_j(\mathbf{R}_c),\\
\boldsymbol{\varphi}_j^{\rm tmp}
&=[\mathbf{r}_j(\mathbf{B}),
\operatorname{Mass}_j(\mathbf{B};\theta_B)],\\
\boldsymbol{\varphi}_j^{\rm cal}
&=[\operatorname{Mass}_j(\mathbf{R}_c;\tau_c),
\mu_j(\mathbf{R}_c\odot\mathbf{B}),\tau_c,\theta_B,\\
&\qquad \operatorname{Top}_{0.20,j}(\mathbf{R}_c)-\tau_c,
\operatorname{Top}_{0.20,j}(\mathbf{B})-\theta_B].
\end{aligned}
\label{eq:supp_region_blocks}
\end{equation}
Their concatenation gives
$\boldsymbol{\varphi}_j=\operatorname{Cat}(\boldsymbol{\varphi}_j^{\rm geo},
\boldsymbol{\varphi}_j^{\rm sem},\boldsymbol{\varphi}_j^{\rm tmp},
\boldsymbol{\varphi}_j^{\rm cal})\in\mathbb{R}^{19}$.
The four blocks encode geometry, semantic confidence, temporal support, and
calibrated margin.

\textbf{Gate Adapter optimization.}
The Gate Adapter predicts
$u_j=\sigma(\mathcal{A}_{\rm gate}(\boldsymbol{\varphi}_j))$.
Source components are labeled from precision- and recall-oriented overlap
with the binary source mask:
\begin{equation}
\begin{aligned}
q_j^{\rm p}
&=\frac{|\omega_j\cap\mathbf{Y}|}{|\omega_j|},\\
q_j^{\rm r}
&=\frac{|\omega_j\cap\mathbf{Y}|}{|\mathbf{Y}|+\epsilon},\\
y_j
&=\mathbf{1}\!\Bigl[
|\omega_j\cap\mathbf{Y}|\ge n_{\min}\\
&\qquad\wedge
(q_j^{\rm p}\ge\delta_p\ \vee\ q_j^{\rm r}\ge\delta_r)
\Bigr],\\
\mathcal{L}_{\rm gate}
&=-\frac{1}{N_\Omega}\sum_j
\Bigl[w_+y_j\log u_j\\
&\qquad +(1-y_j)\log(1-u_j)\Bigr].
\end{aligned}
\label{eq:supp_gate_training}
\end{equation}
Training uses $n_{\min}=8$, $\delta_p=0.30$, $\delta_r=0.05$, and
$w_+=2.0454$. Components outside $[8,262{,}144]$ pixels are excluded from Gate
training. These constants are fixed before target-benchmark evaluation.
During inference, large coherent regions are retained. All other components
must satisfy the learned reliability criterion:
\begin{equation}
\mathbf{M}_c
=\bigcup_{\omega_j\in\Omega_c}
\left\{\omega_j\ \middle|\ 
\frac{|\omega_j|}{HW}\ge\eta_a\ \vee\ u_j\ge\tau_r\right\}.
\label{eq:supp_arf_decision}
\end{equation}
Filtering retains or removes complete components and leaves pixels within each
retained component unchanged.

\subsection{Architecture, Optimization, and Preprocessing}
\label{sec:supp_configuration}

Table~\ref{tab:supp_adapter_architecture} specifies every trainable operator.
The visual encoder, prompt decoder, and CLIP encoders remain frozen. The three
adapters are stored as separate checkpoints.

All optimizers use $\beta_1=0.9$, $\beta_2=0.999$, and
$\epsilon_{\rm opt}=10^{-8}$. Change Adapter training clips the global gradient
norm at 1.0. Images are resized bilinearly to $512\times512$, and masks use
nearest-neighbor resampling. Paired feature jitter uses replicate padding, a
crop displacement of at most four pixels at the highest loaded FPN resolution,
scale-consistent offsets across levels, and horizontal flipping with
probability 0.5. Both timestamps receive the same transform. The Change
Adapter is optimized first; its frozen response archive is then used to build
the memory. The Score and Gate Adapters are fitted on the resulting query- and
component-level representations. All fixed inference values are shared
across target benchmarks: $K_{\rm mem}=16$, $T_m=0.07$,
$[\tau_{\min},\tau_{\max}]=[0.02,0.80]$, correction bound $0.20$, adaptive
floor coefficient $0.75$, $\tau_r=0.60$, $\eta_a=0.04$, and an SCP RMS cap of
$0.25$.

\section{Extended Results}
\label{sec:supp_detailed_results}

This section adds evidence not contained in the main paper. Class-resolved
analyses identify where the aggregate gains occur on SECOND, SCSCD, and xBD.
Additional experiments then quantify query-specific threshold selection and
region-level cue usage, while the qualitative plates expand the number of
examples shown for all seven benchmarks.

\paragraph{Comparison methods.}
We included UCD-SCM \citep{tan2024scm}, AnyChange
\citep{zheng2024anychange}, DynamicEarth \citep{li2026dynamicearth}, UniVCD
\citep{zhu2025univcd}, OmniOVCD \citep{zhang2026omniovcd}, AdaptOVCD
\citep{dou2026adaptovcd}, OpenDPR \citep{guo2026opendpr}, Seg2Change
\citep{su2026seg2change}, and CoRegOVCD \citep{tang2026coregovcd}. Inst-CEG is
the instance-level CEG baseline from SemiCD-VL \citep{li2025semicdvl}.
SAM3-CD denotes the SAM 3 baseline \citep{carion2025sam3} implemented in the
UniVCD study \citep{zhu2025univcd}.
Tables~\ref{tab:supp_second_classwise}--\ref{tab:supp_xbd_classwise} separate
benchmark outcomes from mechanism evidence. Baseline results follow the
benchmark-aligned comparison used in the main paper, and CogVis results are
computed over the complete target test splits using the definitions in
Section~\ref{sec:supp_experimental_details}.
Table~\ref{tab:supp_adapter_evidence} adds query- and region-level measurements
that are not tabulated in the main paper. For each example, every method is
evaluated on the same image pair.

\subsection{Detailed Results on Multi-Class Benchmarks}
\label{sec:supp_classwise_results}

SECOND evaluates direction-invariant class involvement,
\begin{equation}
Y_c^{\rm SECOND}(x)
=\mathbf{1}\!\left[Y^1(x)=c\ \vee\ Y^2(x)=c\right],
\label{eq:supp_second_runtime_target}
\end{equation}
and scores the six foreground masks independently before macro averaging.
SCSCD evaluates seven foreground change categories, while xBD evaluates the
complete five-state damage map. Together, these protocols test whether the
overall gains extend across semantic classes and output spaces.

\paragraph{SECOND.}
Table~\ref{tab:supp_second_classwise} resolves the main-paper aggregate by
foreground category. CogVis ranks first for Building, Tree, Low vegetation,
and Surface and is within 0.02 points of the best Water IoU. These gains span
objects, vegetation, and background-like surfaces rather than one dominant
class.
Playground remains difficult because its mixed impervious and
vegetated appearance increases semantic overlap and fragments connected
support.

\begin{table*}[!t]
\centering
\CogVisCompactTableSetup
\begin{tabular}{@{}lcccccc!{\CogVisDoubleSep}c@{}}
\CogVisTableTopRule
\textbf{Method} & \multicolumn{6}{c}{\textbf{Class-wise IoU / F1}} & \textbf{Mean} \\
\cmidrule(lr){2-7}\cmidrule(l){8-8}
 & \textbf{Building} & \textbf{Tree} & \textbf{Water} & \textbf{Low veg.} & \textbf{Surface} & \textbf{Play.} & \textbf{mIoU / mF1} \\
\CogVisTableHeaderRule
UCD-SCM & 22.06/36.15 & 3.36/6.50 & 1.11/2.20 & 10.35/18.77 & 12.88/22.83 & 0.66/1.31 & 8.40/14.63 \\
AnyChange & 27.22/42.79 & 3.89/7.49 & 1.11/2.19 & 14.17/24.82 & 24.15/38.90 & 0.52/1.03 & 11.84/19.54 \\
Inst-CEG & 23.66/38.27 & 9.64/17.59 & 10.03/18.24 & 0.57/1.14 & 0.00/0.00 & 18.84/31.70 & 10.46/17.82 \\
DynamicEarth-I-M-C & 31.74/48.19 & 10.89/19.64 & 12.34/21.97 & 0.40/0.79 & 0.00/0.00 & 26.93/42.43 & 13.72/22.17 \\
DynamicEarth-M-C-I & 38.60/55.70 & 15.54/26.89 & 15.37/26.64 & 21.25/35.06 & 28.04/43.80 & 22.67/36.97 & 23.58/37.51 \\
SAM3-CD & 22.60/36.90 & 12.40/22.10 & 5.00/9.60 & 18.20/30.90 & 26.70/42.10 & 11.80/21.10 & 16.12/27.12 \\
UniVCD & 43.20/60.40 & 18.90/31.90 & 8.20/15.20 & 24.90/39.80 & 28.00/43.70 & 0.00/0.00 & 20.53/31.83 \\
OmniOVCD & 45.20/62.30 & 16.70/28.60 & 21.20/35.00 & 24.50/39.30 & 27.70/43.40 & 27.00/42.40 & 27.05/41.83 \\
AdaptOVCD & 46.85/63.81 & 12.67/22.49 & 23.34/37.84 & 22.15/36.27 & 33.99/50.74 & 29.28/45.30 & 28.05/42.74 \\
OpenDPR & 42.40/59.50 & 20.90/34.50 & 17.50/29.80 & 23.00/37.40 & 30.20/46.40 & \CogVisBestCell{37.30/54.30} & 28.55/43.65 \\
Seg2Change & \CogVisSecondCell{61.76/76.36} & 16.10/27.73 & 22.88/37.23 & 18.69/31.49 & \CogVisSecondCell{37.26/54.29} & 17.81/30.23 & 29.08/42.89 \\
CoRegOVCD & 48.91/65.69 & \CogVisSecondCell{20.93/34.61} & \CogVisBestCell{29.88/46.01} & \CogVisSecondCell{27.71/43.40} & 32.27/48.79 & \CogVisSecondCell{30.31/46.52} & \CogVisSecondCell{31.67}/\CogVisBestCell{47.50} \\
\CogVisTableSummaryRule
\rowcolor[gray]{0.94}
\textbf{CogVis (Ours)} & \CogVisBestCell{63.07/77.35} & \CogVisBestCell{24.55/39.43} & \CogVisSecondCell{29.86/45.99} & \CogVisBestCell{30.89/47.20} & \CogVisBestCell{39.76/56.90} & 6.73/12.61 & \CogVisBestCell{32.48}/\CogVisSecondCell{46.58} \\
\CogVisTableBottomRule
\end{tabular}
\normalsize
\caption{Class-wise comparison on SECOND. Each entry reports IoU/F1; Mean is the unweighted average across six foreground classes. All values are percentages. Bold and underlined entries indicate the best and second-best results, respectively.}
\label{tab:supp_second_classwise}
\end{table*}

\paragraph{SCSCD.}
Table~\ref{tab:supp_scscd_classwise} shows that CogVis leads on Water and
Vegetation and ranks second on Bareland, Building, Farmland, and Road. Farmland
and Structure exhibit stronger cross-class ambiguity because agricultural
structures, greenhouse-like regions, and built surfaces share closely related
visual patterns.

\begin{table*}[!t]
\centering
\CogVisCompactTableSetup
\begin{tabular}{@{}lccccccc!{\CogVisDoubleSep}c@{}}
\CogVisTableTopRule
\textbf{Method} & \multicolumn{7}{c}{\textbf{Class-wise IoU / F1}} & \textbf{Mean} \\
\cmidrule(lr){2-8}\cmidrule(l){9-9}
 & \textbf{Bare} & \textbf{Water} & \textbf{Building} & \textbf{Struct.} & \textbf{Farm} & \textbf{Veg.} & \textbf{Road} & \textbf{mIoU / mF1} \\
\CogVisTableHeaderRule
UCD-SCM & 8.77/16.13 & 1.69/3.32 & 8.11/15.00 & \CogVisSecondCell{10.09/18.33} & 4.81/9.18 & 9.01/16.52 & 3.06/5.95 & 6.51/12.06 \\
AnyChange & 16.29/28.01 & 2.16/4.22 & 11.03/19.88 & 10.07/18.30 & 5.93/11.19 & 17.85/30.29 & 1.89/3.71 & 9.32/16.51 \\
Inst-CEG & 0.00/0.00 & 7.92/14.68 & 5.45/10.34 & 0.00/0.00 & 0.00/0.00 & 6.79/12.71 & 5.58/10.58 & 3.68/6.90 \\
DynamicEarth-I-M-C & 0.00/0.00 & 12.34/21.97 & 13.55/23.87 & 0.00/0.00 & 0.00/0.00 & 0.00/0.00 & 10.22/18.54 & 5.16/9.19 \\
DynamicEarth-M-C-I & 22.59/36.85 & \CogVisSecondCell{33.14/49.78} & 28.55/44.42 & \CogVisBestCell{13.21/23.33} & 5.48/10.39 & 19.76/33.01 & 3.08/5.97 & 17.97/29.11 \\
Seg2Change & \CogVisBestCell{27.51/43.15} & 21.89/35.92 & \CogVisBestCell{47.20/64.13} & 0.77/1.53 & \CogVisBestCell{11.40/20.47} & \CogVisSecondCell{29.84/45.97} & \CogVisBestCell{23.91/38.60} & \CogVisSecondCell{23.22/35.68} \\
\CogVisTableSummaryRule
\rowcolor[gray]{0.94}
\textbf{CogVis (Ours)} & \CogVisSecondCell{27.28/42.86} & \CogVisBestCell{40.92/58.08} & \CogVisSecondCell{34.33/51.12} & 9.49/17.33 & \CogVisSecondCell{9.35/17.10} & \CogVisBestCell{35.84/52.76} & \CogVisSecondCell{20.33/33.80} & \CogVisBestCell{25.36/39.01} \\
\CogVisTableBottomRule
\end{tabular}
\normalsize
\caption{Class-wise comparison on SCSCD. Each entry reports IoU/F1; Mean is the unweighted average across seven foreground classes. All values are percentages. Bold and underlined entries indicate the best and second-best results, respectively.}
\label{tab:supp_scscd_classwise}
\end{table*}

\paragraph{xBD.}
Table~\ref{tab:supp_xbd_classwise} complements the main-paper mIoU with
state-wise F1 and full-map mF1. CogVis obtains the strongest F1 for no damage,
major damage, and destroyed buildings. The advantage is most pronounced for
major and destroyed structures, where temporal support and region verification
are particularly important. Minor damage remains difficult because its visual
evidence lies between the adjacent no-damage and major-damage states.

\begin{table*}[!t]
\centering
\CogVisCompactTableSetup
\begin{tabular}{@{}lcccc!{\CogVisDoubleSep}cc@{}}
\CogVisTableTopRule
\textbf{Method} & \multicolumn{4}{c}{\textbf{Damage-State F1}} & \multicolumn{2}{c}{\textbf{Full Map}} \\
\cmidrule(lr){2-5}\cmidrule(l){6-7}
 & \textbf{No Damage} & \textbf{Minor} & \textbf{Major} & \textbf{Destroyed} & \textbf{mIoU} & \textbf{mF1} \\
\CogVisTableHeaderRule
Seg2Change & 86.90 & 0.64 & 0.79 & 2.34 & 17.94 & 20.04 \\
OmniOVCD & 87.06 & 1.01 & 0.02 & 9.50 & 18.92 & 21.37 \\
UCD-SCM & 86.94 & 3.07 & 1.86 & 8.75 & 19.04 & 21.62 \\
OpenDPR & 87.31 & \CogVisSecondCell{3.26} & \CogVisSecondCell{3.67} & 12.57 & 19.58 & 22.63 \\
AnyChange & 87.18 & \CogVisBestCell{4.74} & 3.30 & 13.01 & 19.63 & 22.70 \\
AdaptOVCD & \CogVisSecondCell{87.46} & 0.50 & 0.00 & \CogVisSecondCell{39.30} & 22.07 & 26.22 \\
DynamicEarth-M-C-I & 87.41 & 0.01 & 0.00 & 35.42 & \CogVisSecondCell{24.51} & \CogVisSecondCell{29.30} \\
\CogVisTableSummaryRule
\rowcolor[gray]{0.94}
\textbf{CogVis (Ours)} & \CogVisBestCell{88.06} & 0.86 & \CogVisBestCell{11.69} & \CogVisBestCell{50.83} & \CogVisBestCell{32.68} & \CogVisBestCell{40.63} \\
\CogVisTableBottomRule
\end{tabular}
\normalsize
\caption{Class-wise comparison on xBD. The four damage-state columns report F1 on reference building pixels before final background assignment. Full-map mIoU and mF1 are macro-averaged over background and the four damage states after background assignment. All values are percentages; bold and underlined entries indicate the best and second-best results, respectively.}
\label{tab:supp_xbd_classwise}
\end{table*}

\subsection{Additional Calibration and Region Analyses}
\label{sec:supp_controlled_evidence}

To avoid duplicating the analyses already reported in the main paper, this
supplement adds only two complementary measurements. Panel~(a) isolates
query-level threshold selection, and panel~(b) measures how each ARF evidence
family contributes to candidate ranking.

\paragraph{Query-specific threshold calibration.}
We compare a fixed threshold, the deterministic label-free rule, memory
transfer, and the Score Adapter on SECOND and CLCD. The Score Adapter achieves
the highest SECOND mIoU and CLCD IoU (Table~\ref{tab:supp_adapter_evidence},
panel~(a)). Its two-dataset mean exceeds the fixed threshold by 5.94 points and
the memory anchor by 1.81 points, indicating that the learned correction
complements rather than merely reproduces the retrieved operating point.

\paragraph{ARF evidence ablation.}
Removing any evidence family reduces the area under the precision--recall curve
(AUPRC) and the area under the ROC curve (AUROC) on SECOND and CLCD
(Table~\ref{tab:supp_adapter_evidence}, panel~(b)). Temporal support produces
the largest mean decrease, followed by geometry and semantic confidence. These
ranking results indicate that the Gate Adapter uses complementary cues rather
than relying on a single component statistic.

\begin{table*}[!t]
\centering
\CogVisCompactTableSetup
\begin{minipage}[t]{0.48\textwidth}
\centering
\textbf{(a) Query-specific threshold calibration}\par\vspace{0.25em}
\begin{tabular}{@{}L{0.31\linewidth}!{\CogVisDoubleSep}ccc@{}}
\CogVisTableTopRule
\textbf{Threshold rule} & \textbf{SECOND mIoU} & \textbf{CLCD IoU} & \textbf{Mean} \\
\CogVisTableHeaderRule
Fixed global & 23.30 & 15.19 & 19.25 \\
Label-free adaptive & 23.06 & \CogVisSecondCell{17.97} & 20.52 \\
Memory anchor & \CogVisSecondCell{28.81} & 17.94 & \CogVisSecondCell{23.38} \\
Score Adapter & \CogVisBestCell{31.12} & \CogVisBestCell{19.26} & \CogVisBestCell{25.19} \\
\CogVisTableBottomRule
\end{tabular}
\end{minipage}\hfill
\begin{minipage}[t]{0.48\textwidth}
\centering
\textbf{(b) ARF evidence ablation}\par\vspace{0.25em}
\begin{tabular}{@{}L{0.34\linewidth}!{\CogVisDoubleSep}ccc@{}}
\CogVisTableTopRule
\textbf{Metric} & \textbf{Semantic} & \textbf{Temporal} & \textbf{Geometry} \\
\CogVisTableHeaderRule
SECOND $\Delta$AUPRC & $-0.0275$ & $-0.0105$ & $-0.0237$ \\
SECOND $\Delta$AUROC & $-0.0055$ & $-0.0340$ & $-0.0055$ \\
CLCD $\Delta$AUPRC & $-0.0147$ & $-0.0742$ & $-0.0500$ \\
CLCD $\Delta$AUROC & $-0.0056$ & $-0.0887$ & $-0.0033$ \\
Mean drop & $0.0133$ & \CogVisBestCell{$0.0519$} & $0.0206$ \\
\CogVisTableBottomRule
\end{tabular}
\end{minipage}
\normalsize
\caption{Supplementary mechanism evidence not tabulated in the main paper. (a) Query-level threshold calibration; Mean averages SECOND mIoU and CLCD IoU. (b) Region-level ARF evidence ablation under the source-trained overlap criterion. In panel (a), bold and underlined values denote the best and second-best results, respectively; boldface in panel (b) highlights the largest mean drop.}
\label{tab:supp_adapter_evidence}
\end{table*}

\subsection{Qualitative Comparison}
\label{sec:supp_qualitative}

\begin{figure*}[p]
\centering
\includegraphics[width=0.95\textwidth]{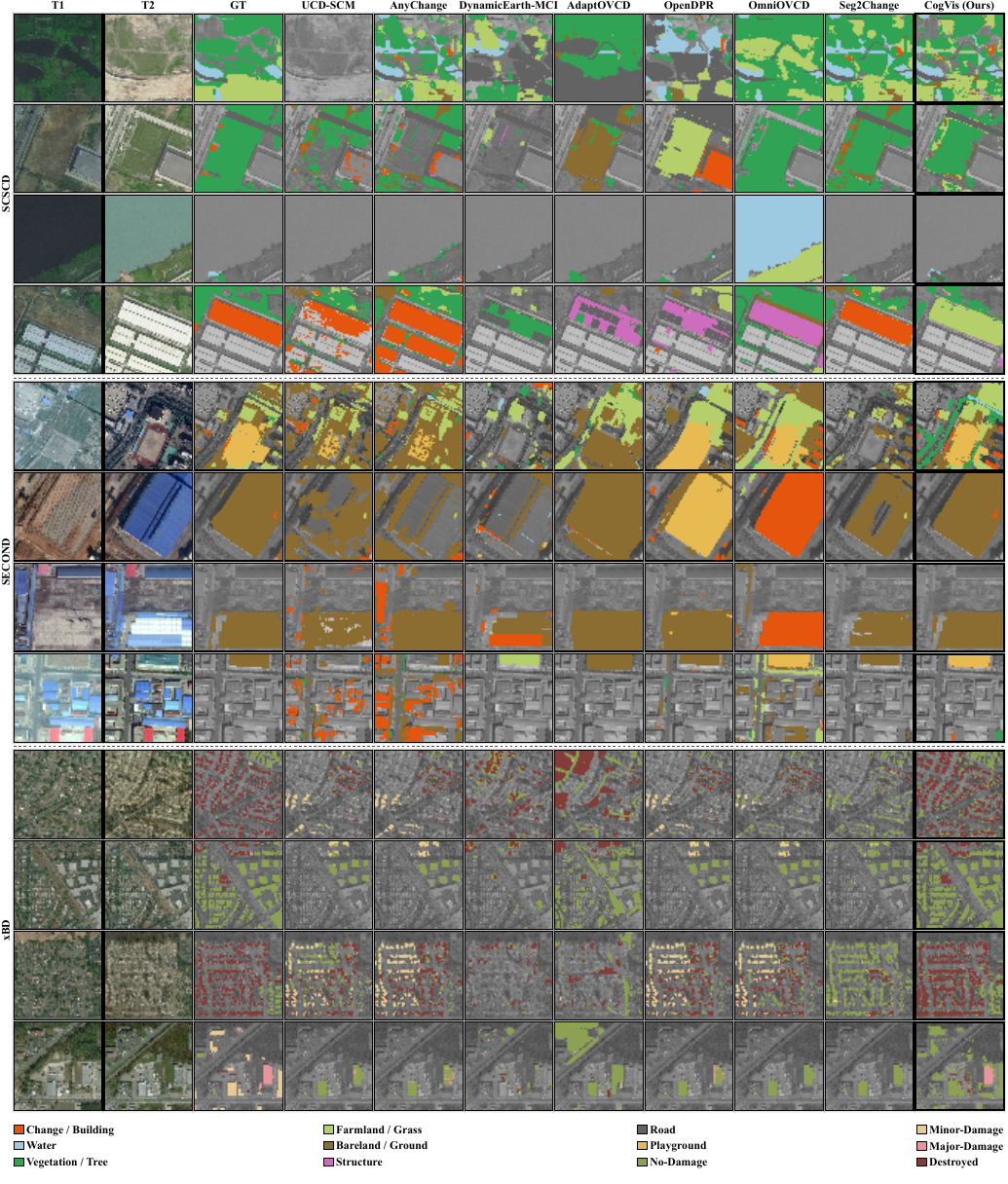}
\caption{Qualitative comparison on SCSCD, SECOND, and xBD. Four examples are shown per dataset. Columns show $T_1$, $T_2$, ground truth, seven baselines, and CogVis; colors denote semantic change classes on SCSCD and SECOND and damage states on xBD.}
\label{fig:supp_multiclass_comparison}
\end{figure*}

\begin{figure*}[p]
\centering
\includegraphics[width=0.98\textwidth]{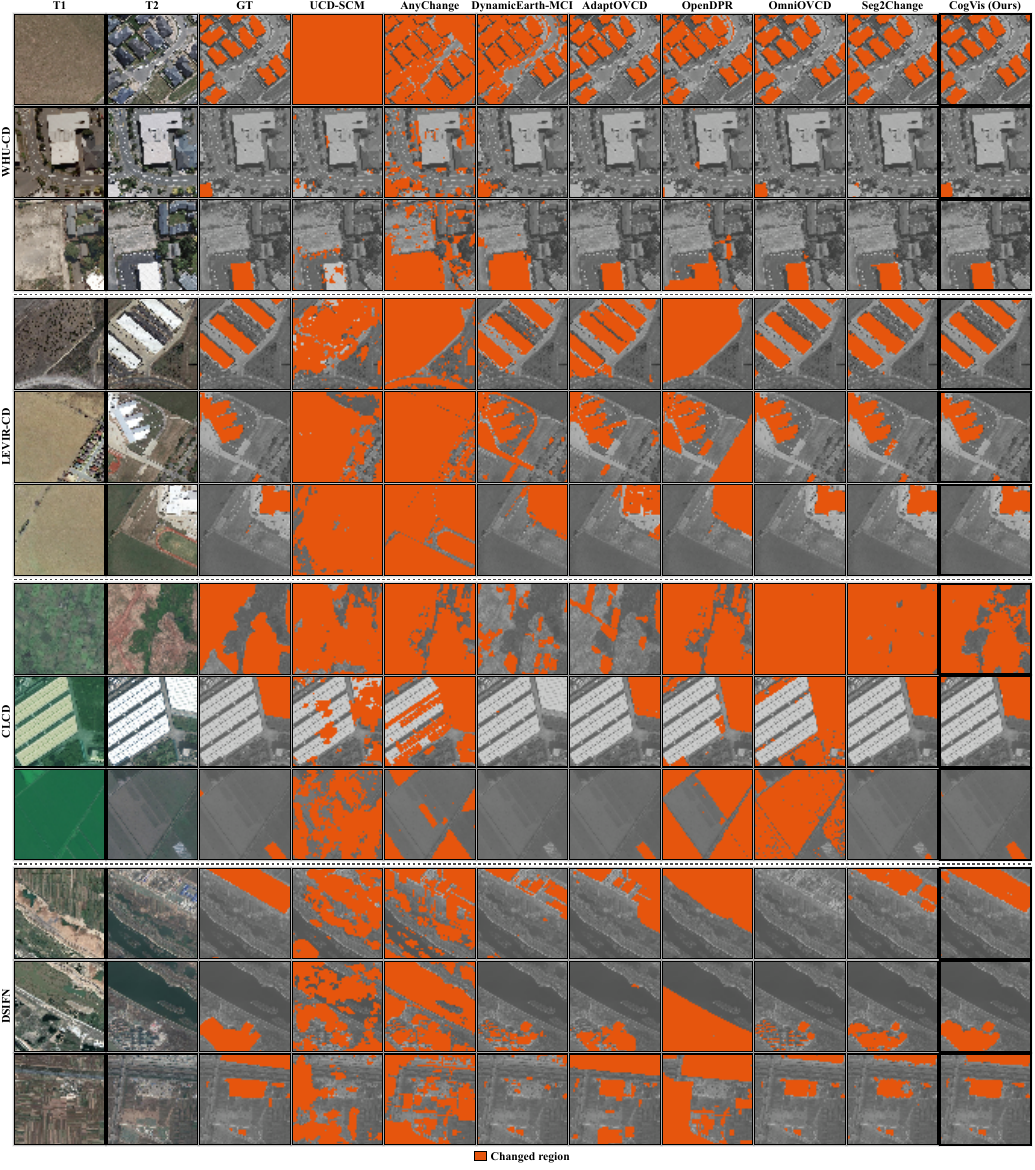}
\caption{Qualitative comparison on WHU-CD, LEVIR-CD, CLCD, and DSIFN. Three examples are shown per dataset. Columns show $T_1$, $T_2$, ground truth, seven baselines, and CogVis; orange denotes changed pixels overlaid on $T_2$.}
\label{fig:supp_binary_comparison}
\end{figure*}

The main paper provides a consolidated qualitative overview.
Figures~\ref{fig:supp_multiclass_comparison} and
\ref{fig:supp_binary_comparison} expand that evidence with four pairs from
each of SECOND, SCSCD, and xBD and three pairs from each of WHU-CD, LEVIR-CD,
CLCD, and DSIFN. For each example, every method is evaluated on the same image
pair; the additional samples expose category-specific and nuisance-driven
failure patterns rather than repeat the main-paper examples.

SECOND and SCSCD are rendered with a symmetric class-involvement target
\begin{equation}
M_c^{\rm vis}=M_{\rm CD}\wedge[(Y^1=c)\vee(Y^2=c)].
\label{eq:supp_visual_target}
\end{equation}
UCD-SCM and AnyChange produce category-agnostic masks. A common frozen
open-vocabulary classifier therefore assigns semantic labels within their
predicted support. Native semantic outputs are rendered directly.

Across the semantic examples, CogVis confines Building, Water, Vegetation, and
Surface responses to object-aligned regions while preserving their class
identity. In the binary panel, its clearest visible advantage is the rejection
of broad false-positive masks that saturate stable roofs, fields, or background
in several category-agnostic outputs. The retained responses instead follow
changed structures in the post-event image. On xBD, building support and
severe-damage regions are delineated more coherently, consistent with the
full-map gains in Table~\ref{tab:supp_xbd_classwise}. These displayed patterns
are consistent with the combined effects of pair-level suppression,
query-specific calibration, and component-level filtering.

\subsection{Category-Level Interpretation}
\label{sec:supp_failure_modes}
The remaining errors concentrate in categories with mixed visual composition
or fine-grained semantic transitions. Playground combines impervious surfaces
with vegetation, while Structure overlaps visually with buildings and
agricultural structures. Damage grades with subtle appearance differences
require finer separation than the endpoint states. These patterns indicate
that the remaining performance gap is primarily associated with fine-grained
semantic discrimination. In contrast, the consistent gains on Building,
Water, Vegetation, Surface, and the complete xBD map demonstrate that the
scope-aligned design remains effective across diverse spatial structures and
change types.

\section{Large-Scene Wildfire Damage Mapping}
\label{sec:supp_wildfire_case}

\paragraph{Application setting.}
We mapped registered pre- and post-event RGB imagery from the January 2025
Palisades and Eaton fires in Los Angeles County \citep{seydi2025wildfires}
using the trained CogVis model and xBD prompt vocabulary without site-specific
fine-tuning. The aligned Palisades pair contains
$17{,}592\times13{,}442$ pixels at approximately $0.51$~m ground sampling
distance and covers $8.98\times6.86$~km. The Eaton pair contains
$12{,}721\times10{,}455$ pixels at approximately $0.55$~m ground sampling
distance and covers $7.01\times5.76$~km. Both pairs use EPSG:32611 and a shared
within-scene affine transform. Because site-specific reference labels are
unavailable, this case study supports only qualitative spatial transfer, not
local accuracy or damage-extent estimates.

\paragraph{Mapping procedure.}
We partitioned each scene into $512\times512$ tiles at a 384-pixel stride,
yielding 1,610 Palisades tiles and 891 Eaton tiles. For tile prediction
$\widehat{y}_t(x)$ and class $c$, overlap was fused by
\begin{equation}
\begin{aligned}
V_c(x)
&=\frac{\sum_{t\ni x}w_t(x)\,
\mathbf{1}[\widehat{y}_t(x)=c]}
{\sum_{t\ni x}w_t(x)},\\
\widehat{y}(x)
&=\arg\max_c V_c(x),
\end{aligned}
\label{eq:supp_wildfire_fusion}
\end{equation}
where $w_t$ is a separable two-dimensional Hann window with a $10^{-3}$ border
floor. The weighted vote suppresses tile-edge discontinuities while retaining
a defined estimate at the scene boundary.

\begin{figure*}[t]
\centering
\includegraphics[width=\textwidth]{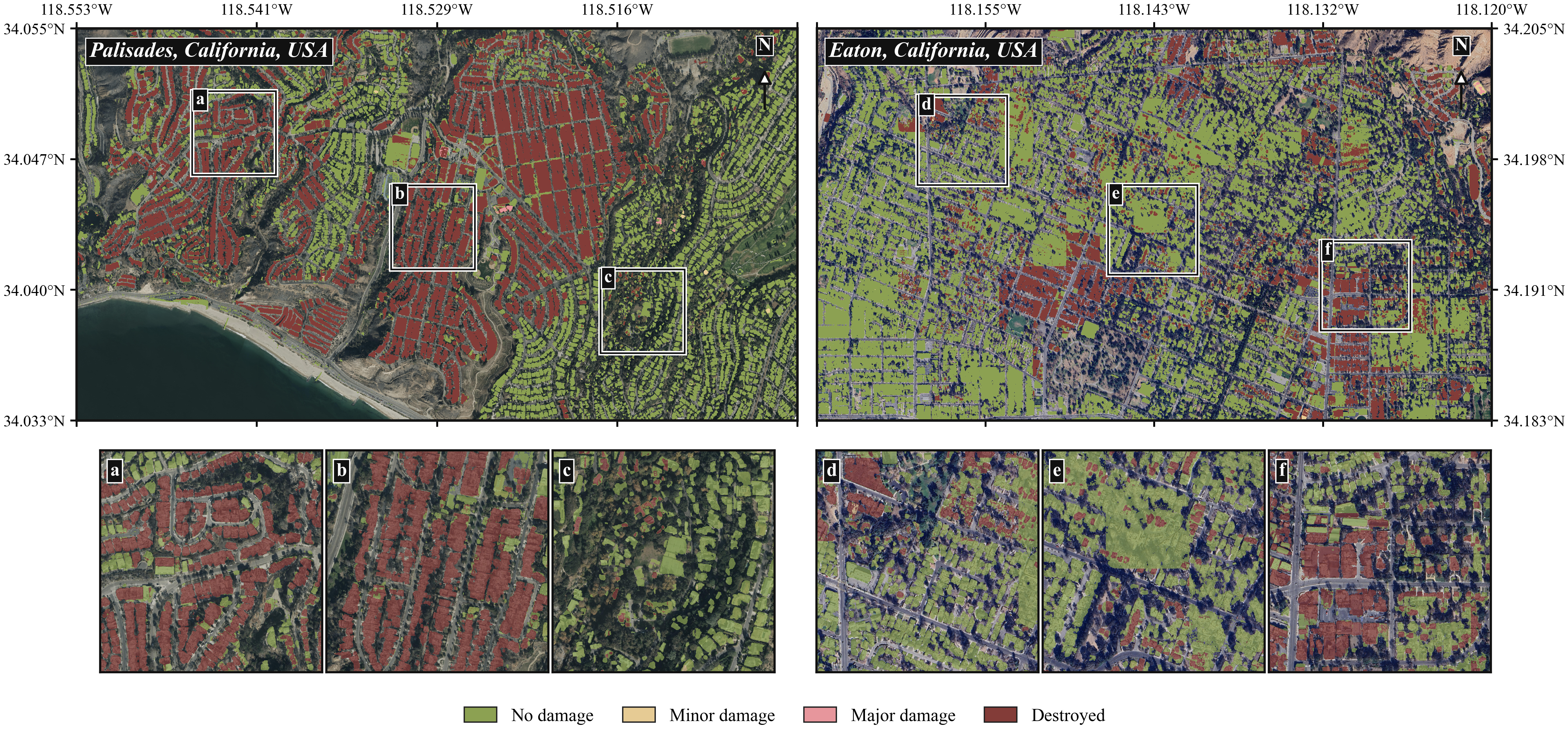}
\caption{Large-scene wildfire damage predictions for the Palisades and Eaton fire areas in California. Post-event overview maps show Palisades (left) and Eaton (right), with $1024\times1024$-pixel regions (a)--(c) and (d)--(f) enlarged below. Detail panels overlay predictions on RGB imagery. Olive green, tan, rose, and dark red denote no damage, minor damage, major damage, and destroyed, respectively; axis ticks report WGS~84 coordinates, and arrows indicate north.}
\label{fig:supp_wildfire_case}
\end{figure*}

\paragraph{Geospatial sampling and rendering.}
Figure~\ref{fig:supp_wildfire_case} overlays the fused maps on post-event RGB
imagery. Three $1024\times1024$-pixel regions were sampled at fixed relative
positions in each scene and grid-adjusted to retain at least 99.9\% paired
coverage. Panels~(a)--(c) show Palisades, and panels~(d)--(f) show Eaton.
Overview maps use opaque class colors; detail panels use 0.72 opacity to retain
roof, parcel, and street context.

\paragraph{Prompt formulation.}
The standalone phrases in Table~\ref{tab:supp_damage_prompts} were applied
verbatim to both scenes. Five building-support phrases localized candidate
structures; four state-specific banks distinguished no damage, minor damage,
major damage, and destroyed. Phrases were encoded independently, and responses
within each state were aggregated by pixelwise maximum, preserving the
vocabulary and inference rule evaluated on xBD.

\paragraph{Spatial patterns.}
Palisades contains broad, contiguous destroyed-state responses. Panels~(a)
and~(b) resolve these responses along roof rows and residential blocks;
panel~(c) instead shows a predominantly no-damage area interspersed with smaller
destroyed regions. Eaton is dominated by no-damage responses. Panels~(d)
and~(e) isolate smaller destroyed clusters within the urban fabric, whereas
panel~(f) contains the most compact Eaton response. Across 2,501 tiles, the
fused predictions preserve roof- and block-level spatial continuity without
obvious seams at the displayed scale. These maps show how the frozen model and
prompt vocabulary can be applied to large geospatial scenes through
overlap-aware tiled inference. Figure~\ref{fig:supp_wildfire_case} provides a
scene-scale visualization of the same fixed inference rule evaluated on xBD.

\end{document}